\pdfoutput=1

\documentclass[11pt]{article}

\usepackage[preprint]{acl}

\usepackage{times}
\usepackage{latexsym}
\usepackage{colortbl}
\usepackage{balance}
\usepackage[T1]{fontenc}

\usepackage[utf8]{inputenc}

\usepackage{microtype}

\usepackage{inconsolata}
\usepackage{scalerel}
\usepackage{booktabs}
\usepackage{multirow}
\usepackage{graphicx}
\usepackage{titlesec}
\usepackage{amsmath}
\usepackage{subcaption}
\usepackage{colortbl}  

\newcommand{\ourdata}{\texttt{INTERS}}
\newcommand{\ie}{\textit{i.e.}}
\newcommand{\eg}{\textit{e.g.}}

\title{
\raisebox{-0.35cm}{\includegraphics[height=1.2cm]{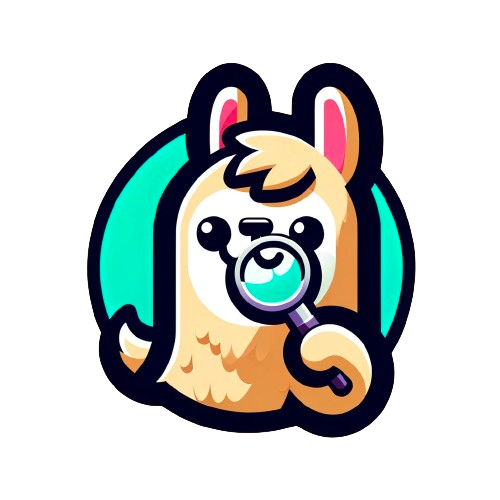}} \texttt{INTERS}: Unlocking the Power of Large Language Models in Search \\ with Instruction Tuning
}

\author{Yutao Zhu$^1$, Peitian Zhang$^1$, Chenghao Zhang$^{1,2*}$, Yifei Chen$^{1,3*}$, Binyu Xie$^1$ \\ \textbf{Zheng Liu}$^4$, \textbf{Ji-Rong Wen}$^1$, \and \textbf{Zhicheng Dou}$^{1\dag}$ \\
$^1$Gaoling School of Artificial Intelligence, Renmin University of China \\
$^2$School of Computer Science, Beijing University of Posts and Telecommunications \\
$^3$School of Artificial Intelligence, Nankai University, $^4$Beijing Academy of Artificial Intelligence \\
\texttt{yutaozhu94@gmail.com, dou@ruc.edu.cn}
}

\begin{document}
\maketitle
\def\thefootnote{*}\footnotetext{This work was done when Chenghao Zhang and Yifei Chen were doing internships at Renmin University of China.}
\def\thefootnote{\dag}\footnotetext{Corresponding author.}
\def\thefootnote{\arabic{footnote}}

\begin{abstract}

Large language models (LLMs) have demonstrated impressive capabilities in various natural language processing tasks. Despite this, their application to information retrieval (IR) tasks is still challenging due to the infrequent occurrence of many IR-specific concepts in natural language. While prompt-based methods can provide task descriptions to LLMs, they often fall short in facilitating a comprehensive understanding and execution of IR tasks, thereby limiting LLMs' applicability. To address this gap, in this work, we explore the potential of instruction tuning to enhance LLMs' proficiency in IR tasks. We introduce a novel instruction tuning dataset, \ourdata{}, encompassing 20 tasks across three fundamental IR categories: query understanding, document understanding, and query-document relationship understanding. The data are derived from 43 distinct datasets with manually written templates. Our empirical results reveal that \ourdata{} significantly boosts the performance of various publicly available LLMs, such as LLaMA, Mistral, and Falcon, in IR tasks. Furthermore, we conduct extensive experiments to analyze the effects of instruction design, template diversity, few-shot demonstrations, and the volume of instructions on performance. We make our dataset and the fine-tuned models publicly accessible at~\url{https://github.com/DaoD/INTERS}.
\end{abstract}

\section{Introduction}
Large language models (LLMs) have shown remarkable capabilities across various natural language processing (NLP) tasks. While these models have learned vast knowledge from large text corpora, their (pre-)training objective is not aligned with human's objective: the latter requires models to ``follow human instructions and perform tasks'' rather than ``predict the next token''. To address this mismatch, instruction tuning is proposed, serving as an effective technique to align LLMs with human tasks and preferences~\cite{instruct-gpt,flan,flanv2,natural-instructions,super-natural-instructions,self-instruct}. After instruction tuning, LLMs can better understand users' intent and show impressive generalization to new tasks.

In the area of information retrieval (IR), the introduction of LLMs has also led to notable developments~\cite{wang2023query2doc,PromptCase,rankgpt,rankllama}. Due to the high cost of fine-tuning, many existing studies leverage prompting methods to apply LLMs in IR tasks. However, some of them have reported that LLMs cannot consistently outperform fine-tuned smaller models in this manner~\cite{rankgpt,hyde}. For example, RankGPT~\cite{rankgpt} based on \texttt{gpt-3.5-turbo} underperforms monoBERT with 340M parameters on passage ranking tasks. This discrepancy may stem from the complexity of IR-specific concepts like queries, relevance, and search intent, which are infrequently encountered in pre-training corpora and are inherently challenging to comprehend. 


\begin{figure}
    \centering
    \includegraphics[width=\linewidth]{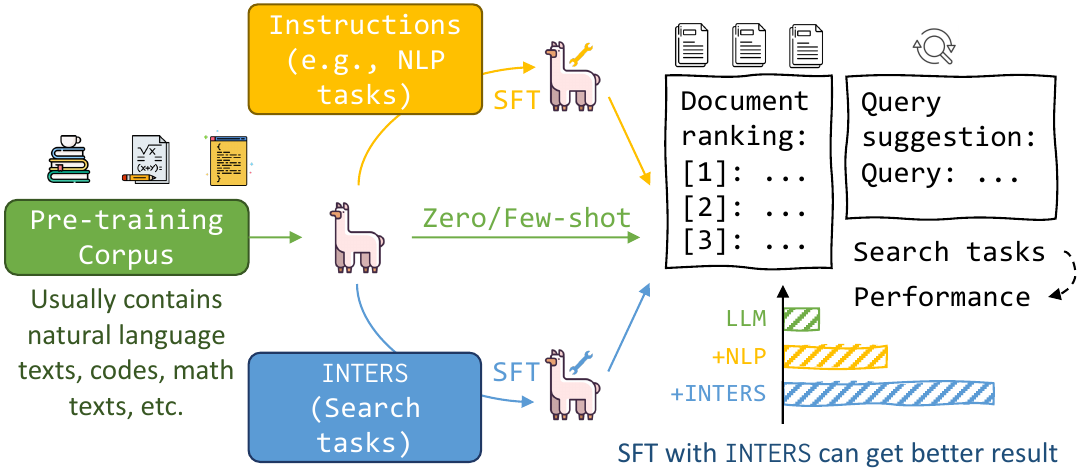}
    \caption{Compared with existing datasets, \ourdata{} is designed specifically for search tasks.}
    \label{fig:intro}
\end{figure}

To fill the gap, in this work, we build a novel \textbf{IN}struction \textbf{T}uning datas\textbf{E}t fo\textbf{R} \textbf{S}earch (\ourdata{}).\footnote{This paper will use the terms ``search-related tasks'' and information retrieval tasks'' interchangeably.} This dataset is designed to specifically enhance the search capabilities of LLMs. Since the search process involves various tasks, we choose to focus on three pivotal aspects: query understanding, document understanding, and the understanding of the relationship between queries and documents. We collect 43 datasets covering 20 distinct tasks, ensuring the dataset's comprehensive coverage and \textit{richness}. In order to improve \textit{diversity} and broaden applicability, we manually craft 12 unique templates for each dataset and consider both zero-shot and few-shot examples in data generation. To further improve the models' \textit{generalizability}, we also manually write a detailed task description for each task, which serves as a bridge to connect each dataset under the same task. Finally, we plan to release all data, templates, model checkpoints, experimental results, and source codes. We wish the \textit{openness} of our dataset could support more future research and development in this field.

We conduct experiments by fine-tuning several open-sourced LLMs using the \ourdata{} dataset. Experimental results show that \ourdata{} consistently enhances the performance of LLMs of different sizes across a spectrum of search tasks. Notably, this improvement is observed not only in tasks that are directly learned in the training data (in-domain) but also in tasks that are unseen in the training set (out-of-domain). Our further experiments highlight several key insights: (1) customized templates and task descriptions effectively improve model performance; (2) the diversity of templates can enhance model generalizability; (3) instruction tuning specifically tailored for search tasks addresses the existing gap of NLP instructions for such tasks; (4) combining instruction tuning and few-shot prompting can further improve performance; and (5) the substantial data volume can benefit the efficacy of instruction tuning.

\section{Related Work}

\paragraph{Large Language Models for Information Retrieval}
LLMs possess a remarkable capacity for language understanding, enabling them to be highly valuable in comprehending user queries and documents. Therefore, many researchers have explored applying LLMs to IR tasks~\cite{llm4ir}. Existing studies can be roughly categorized into two groups. The first group of studies treats LLMs as search agents to accomplish search tasks~\cite{webgpt,webcpm,webglm}. A typical method is WebGPT~\cite{webgpt}, which employs imitation learning to teach an LLM (\ie, GPT-3) to use search engines and answer questions like a human. The other group of studies mainly focuses on applying LLMs to specific IR tasks, such as query reformulation~\cite{wang2023query2doc,srinivasan2022quill,PromptCase,DBLP:conf/emnlp/MaoDMH0Q23} and document ranking~\cite{rankgpt,rankinggpt,rankllama,opensource}. Most of these studies rely on prompting LLMs in a zero-shot or few-shot manner. However, due to the inherent complexity of the IR task and the relative scarcity of IR-related concepts in natural language texts, LLMs often cannot achieve superior performance to fine-tuned smaller models in IR tasks~\cite{rankgpt,hyde}.



Different from existing studies, our research focuses on using instruction tuning to improve the overall performance of LLMs on various search tasks. This involves enhancing the models' abilities to interpret and respond to search-related instructions more effectively, thereby improving their utility in complex IR scenarios.

\paragraph{Instruction Tuning for LLMs}
Instruction tuning (IT) aims at fine-tuning pre-trained LLMs on a collection of formatted instances in the form of natural language~\cite{flan,natural-instructions,super-natural-instructions,self-instruct}. After IT, LLMs can better follow instructions and perform human tasks. This approach bears a close resemblance to supervised fine-tuning~\cite{instruct-gpt} and multi-task prompt training~\cite{t0}. Instruction tuning's efficacy lies in its ability to not only enhance LLMs' performance on tasks they have been directly trained on but also to equip them with the ability to generalize to new, unseen tasks~\cite{t0,flan}. 

In this work, we leverage IT to specifically enhance LLMs' performance on search-related tasks. Our dataset is designed with a deep understanding of the task characteristics. Experiments will show that IT is also an effective way to improve LLMs' overall performance on search tasks.

\begin{figure*}
    \centering
    \includegraphics[width=.8\linewidth]{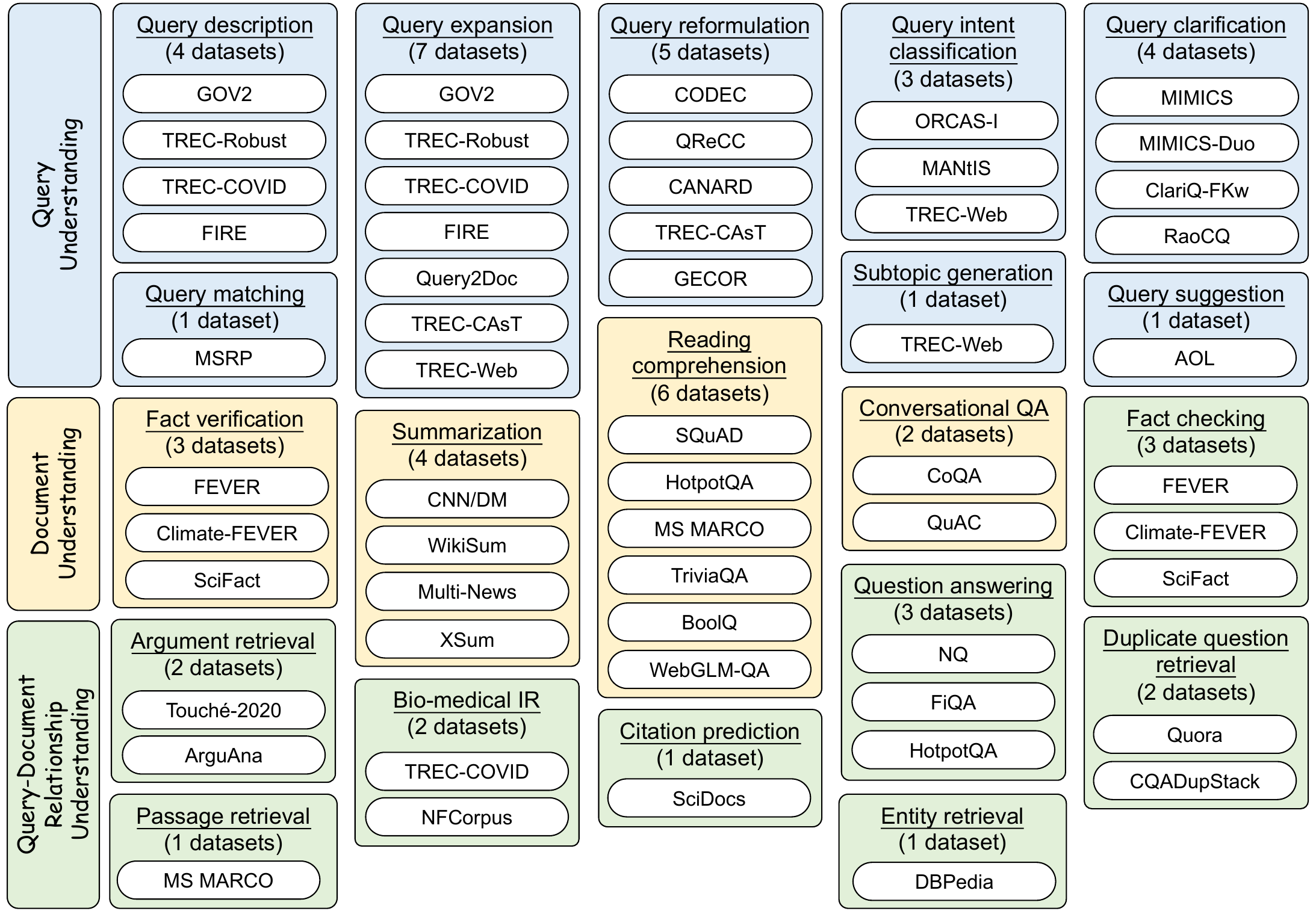}
    \caption{Categories, tasks, and datasets used in \ourdata{}. Different colors indicate different task categories.}
    \label{fig:datasets}
\end{figure*}

\section{Instruction Tuning for Search}
Instruction tuning has proven to be effective for LLMs in responding to instructions. This method essentially involves training LLMs through supervised learning to execute particular tasks based on provided instructions. A notable benefit of this approach is that, after fine-tuning, LLMs can comprehend and execute instructions not only for similar tasks but also for tasks they have not learned before. However, it is important to note that search tasks, which are the focus of our study, differ significantly from typical NLP tasks in terms of their objectives and structures. Search tasks primarily revolve around two key elements: \textit{queries} and \textit{documents}. Therefore, as shown in Figure~\ref{fig:datasets}, we consider collecting tasks and datasets in three categories: query understanding, document understanding, and query-document relationship understanding. We consider that tasks within these categories are instrumental in refining LLMs’ abilities to interpret queries, comprehend documents, and understand their relationships. Below, we introduce the tasks, datasets, and data construction process.

\subsection{Tasks \& Datasets}
Developing a comprehensive instruction-tuning dataset covering a wide range of tasks is very resource-intensive. To address this, we follow the previous studies~\cite{flan,flanv2} and choose to convert existing datasets from the IR research community into an instructional format. We consider tasks under the categories of query understanding, document understanding, and query-document understanding. All tasks and datasets we used are shown in Figure~\ref{fig:datasets}. Their detailed descriptions, evaluation metrics, licenses, and examples are provided in Appendix~\ref{app:task} and~\ref{app:example}. 

\paragraph{Query Understanding}
In IR, a query is a user-initiated request for information, typically composed of keywords, phrases, or natural language questions. It aims at retrieving relevant information from a retrieval system (\eg, a search engine). The effectiveness of a query is measured by its ability to accurately reflect the user's intent and retrieve the most relevant documents. During the retrieval process, query understanding is a critical component in determining the efficiency and user satisfaction of the IR systems. Therefore, we collect a group of tasks (eight in total) that require models to understand the semantics of queries and capture the underlying user search intent.

\paragraph{Document Understanding}
In IR, a document refers to any piece of information that can be retrieved in response to a query, such as web pages in search engines. Document understanding is the process by which an IR system interprets and comprehends the content of these documents. Enhanced document understanding leads to better search results and an overall more efficient and user-friendly retrieval process. 
In \ourdata{}, we collect datasets for four tasks that require a deep understanding of documents.

\paragraph{Query-document Relationship Understanding}
Query-document relationship understanding is the process of determining how well the content of a document matches or satisfies the intent behind a user's query. This involves interpreting the query's semantics, context, and purpose, and then assessing the relevance of documents based on how closely they correspond to these aspects. It is the core task of information retrieval. We collect eight tasks specifically designed to enhance models' capability of determining various query-document relationships, \eg, the question answering task involves understanding the relationship between questions and supporting evidences. It is important to recognize the variety of architectures available for modeling the query-document relationship. In this research, we focus on the reranking architecture, which is the most straightforward way to apply LLMs. The candidate documents for reranking are retrieved by BM25~\cite{bm25}.


\begin{figure*}[t]
    \centering
    \includegraphics[width=\linewidth]{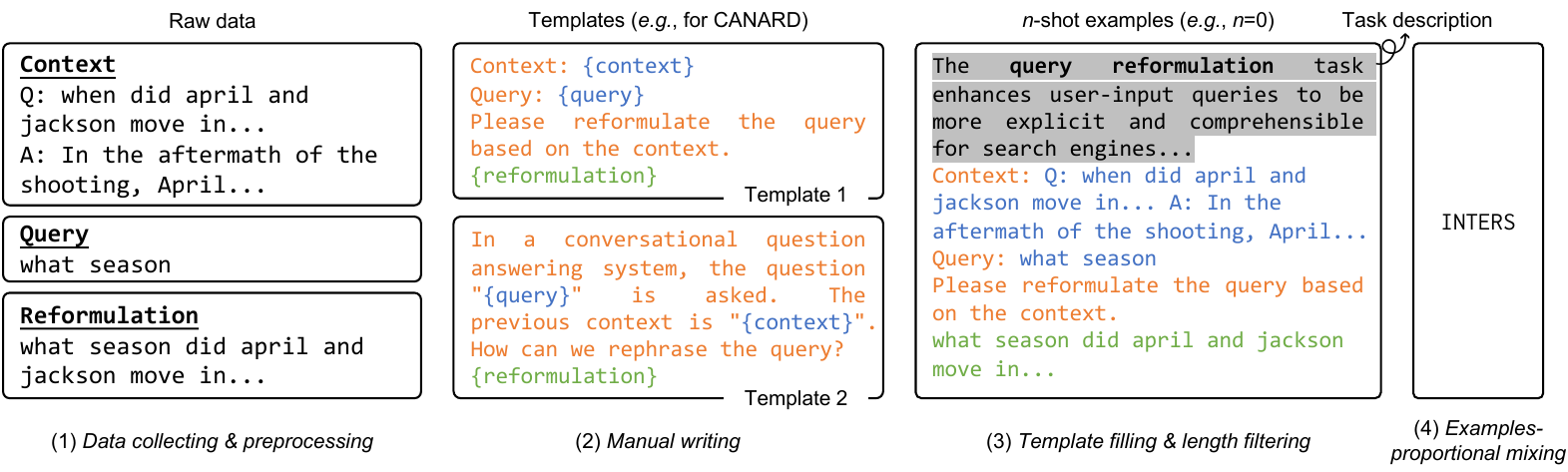}
    \caption{An example of our data construction process.}
    \label{fig:construction}
\end{figure*}

\subsection{\ourdata{} Construction}\label{sec:construction}
After determining the tasks and datasets we plan to use, we start to construct \ourdata{}. The construction process is illustrated in Figure~\ref{fig:construction}, which can be divided into four steps.

(1) \textbf{Preprocessing}. We download all datasets from publicly available resources, filter out unnecessary attributes and invalid data samples, and then convert them into the JSONL format for further processing. 

(2) \textbf{Template collection}. We manually craft 12 distinct templates for \textit{each dataset} to ensure the diversity and richness of the generated data. These templates use natural language instructions to describe the specific task associated with each dataset (two example templates are shown in the second part of Figure~\ref{fig:construction}). To further improve the diversity of the templates, following the design of FLAN~\cite{flan}, we integrate up to two ``inverse'' templates per dataset. For example, for the query expansion task, we include templates that prompt for simplifying a query. In particular, for query-document relationship understanding tasks, there are three typical methods~\cite{llm4ir, DBLP:journals/corr/abs-2306-17563}, \ie, pointwise, pairwise, and listwise. We consider all of them when writing templates (\ie, four for each type) to support a wider range of application scenarios (related discussion is presented in Section~\ref{sec:ranking}).
Additionally, to enhance the LLMs' task comprehension, we provide detailed descriptions for \textit{each task}. These task descriptions serve a dual purpose: offering a granular understanding of the task's objectives and establishing a linkage among datasets under the same task. The efficacy of this design will be demonstrated through our experiments presented in Section~\ref{sec:task-description}.

(3) \textbf{Example generation}. For each sample in the preprocessed data, we use the corresponding task description and a randomly selected template to generate $n$-shot examples (where $n\in[0, 5]$ in our experiments). The third part of Figure~\ref{fig:construction} shows a zero-shot example generated from the CANARD dataset. For few-shot examples (where $n\geq 1$), we insert the $n$ examples between the task description and the input, where the examples are separated by special tokens (\ie, ``\textbackslash n\textbackslash n''). All few-shot examples are randomly selected from the training set. 
Moreover, to ensure that the few-shot examples are within the learnable scope of LLMs, we apply a length filter to exclude examples that exceed a predefined length threshold (2,048 tokens in our experiments). 

(4) \textbf{Example mixture}. To compile \ourdata{}, we randomly select examples from our entire collection until we accumulate a total of 200k examples.\footnote{This number is determined to strike a balance between efficacy and training costs.} To balance the different sizes of datasets, 
we apply an examples-proportional mixing strategy~\cite{t5} with a mixing rate maximum of 5k. Under this scheme, any dataset contributing more than 5k examples does not receive extra weighting for the additional samples, thus preventing the dominant influence from larger datasets.

\section{Experiments}
We fine-tune several open-sourced LLMs on our \ourdata{}, and evaluate their performance in different settings. Our experiments will investigate the following research questions:
(1) Can the model obtain the capability to solve search tasks through instruction tuning on \ourdata{}? ({$\S$\ref{sec:in-domain-eval}})
(2) Is this capability generalizable? ({$\S$\ref{sec:out-of-domain-eval}})
(3) Is our instruction design effective?
({$\S$\ref{sec:task-description}})
(4) Are there any advantages compared to existing instruction sets? ($\S$\ref{sec:flan})
(5) Is the model still effective with few-shot demonstrations?  ($\S$\ref{sec:few-shot})
(6) What is the impact of data volume? ($\S$\ref{sec:data-volume})
(7) What are the effects of different ranking strategies? ($\S$\ref{sec:ranking})

\subsection{Backbone Models}
We consider four LLMs in different sizes, ranging from 1B parameters to 7B parameters: \textbf{Falcon-RW-1B}~\cite{falcon-rw-1b}, \textbf{Minima-2-3B}~\cite{minima-2-3b}, \textbf{Mistral-7B}~\cite{mistral-7b}, and \textbf{LLaMA-2-7B}~\cite{llama2}. These models are publicly available and widely used in many studies. 
Their detailed introduction and implementation details are provided in Appendix~\ref{app:implementation}.

\begin{figure*}[t]
    \centering
    \includegraphics[width=\linewidth]{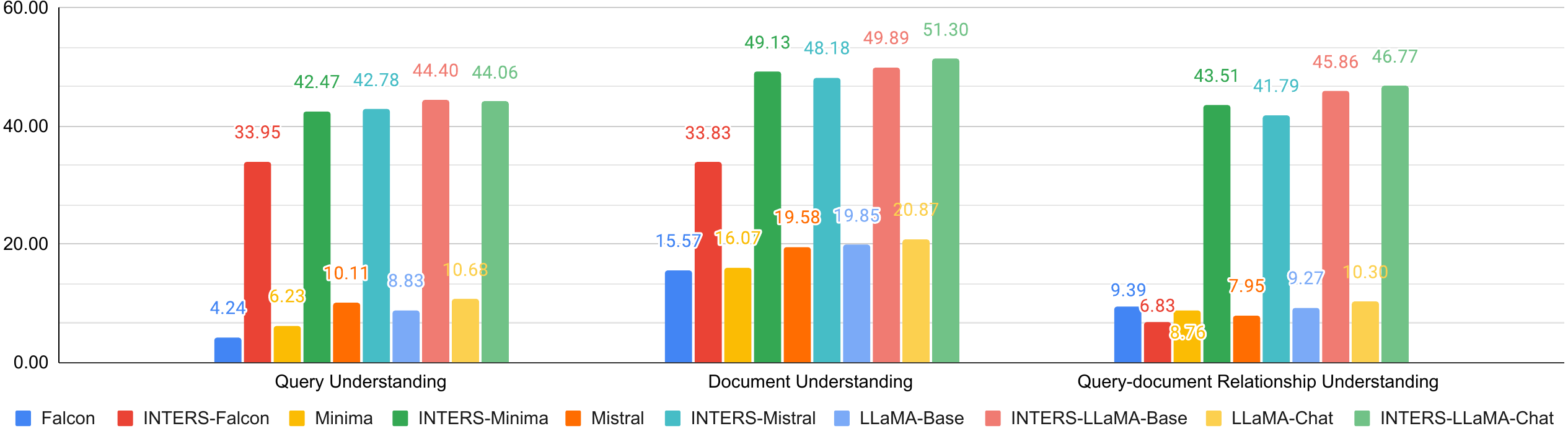}
    \caption{Average performance of all models and fine-tuned models under zero-shot settings. For query-document relationship understanding tasks, we use pointwise methods. The full results are shown in Appendix~\ref{app:add_result}.}
    \label{fig:in-domain}
\end{figure*}

\subsection{In-domain Evaluation}\label{sec:in-domain-eval}
We first perform an in-domain evaluation to validate the effectiveness of instruction tuning with \ourdata{} on search tasks. In this experiment, we split all data into training, validation, and test sets (details are presented in Appendix~\ref{app:eval}). The models are fine-tuned on the training set and evaluated on the test set. As all tasks and datasets are exposed during training, we call it an in-domain evaluation. 

The experimental results are shown in Figure~\ref{fig:in-domain}. Generally, after fine-tuning on \ourdata{}, all models of varying sizes can achieve significantly better performance, demonstrating the effectiveness and broad applicability of instruction tuning in enhancing LLMs' search performance. Besides, we have the following observations. 

(1) On most datasets, larger models tend to perform better than smaller ones. For instance, LLaMA (7B) and Mistral (7B) show superior performance compared to Minima (3B) and Falcon (1B). Intriguingly, in query-document relationship understanding tasks, larger models without fine-tuning can even outperform the smaller models after fine-tuning (\eg, LLaMA-Chat > INTERS-Falcon). This confirms the inherent advantages of larger-scale parameters in model performance. (2) We notice that INTERS-Falcon exhibits inferior performance compared to untuned Falcon in tasks related to understanding the query-document relationship. We attribute this to the complex nature of these tasks. (3) Notably, in document understanding tasks, INTERS-Minima (3B) outperforms INTERS-Mistral (7B), suggesting that fine-tuning smaller models could serve as a cost-effective approach for particular tasks.
(4) Before fine-tuning, LLaMA-Chat, which has already been optimized for dialogue scenarios, exhibits superior performance compared to LLaMA-Base. This advantage is attributed to LLaMA-Chat's better capability of understanding instructions and performing tasks. However, after instruction tuning with \ourdata{}, the performance gap diminishes. This shows the broad generality of our instruction tuning for various types of LLMs.

\begin{table}[t]
    \centering
    \small
    \setlength{\tabcolsep}{1.8mm}{
    \begin{tabular}{lccccc}
    \toprule
        Category & No FT & \ourdata{} & \textit{w/o} Q & \textit{w/o} D & \textit{w/o} Q-D  \\
    \midrule
        Q & $10.68$ & $44.06$ & $15.35$ & $43.11$ & $43.76$ \\
        D & $20.87$ & $51.30$ & $51.05$ & $21.09$ & $51.11$ \\
        Q-D & $10.30$ & $46.77$ & $47.40$ & $45.99$ & $29.36$ \\
        Avg. & $13.18$ & $46.42$ & $33.20$ & $37.58$ & $41.75$ \\
    \bottomrule
    \end{tabular}
    }
    \caption{Average performance of removing different task categories. ``Q'', ``D'', and ``Q-D'' denote query understanding, document understanding, and query-document relationship understanding, respectively.}
    \label{tab:remove_group}
\end{table}

\begin{table}[t]
    \centering
    \small
    \setlength{\tabcolsep}{1.7mm}{
    \begin{tabular}{lcccccc}
    \toprule
        Task & No FT & \ourdata{} & \textit{w/o} QIC &\textit{w/o} FV & \textit{w/o} CP \\
    \midrule
        QIC & $20.32$ & $53.92$ & $38.55$ & $50.72$ & $50.08$ \\
        QR & $7.82$ & $69.39$ & $68.69$ & $69.69$ & $68.77$ \\
        FV & $48.75$ & $76.29$ & $75.09$ & $49.08$ & $76.10$ \\
        Summ. & $11.11$ & $20.98$ & $20.27$ & $21.30$ & $21.37$ \\
        CP & $2.90$ & $16.71$ & $18.02$ & $18.66$ & $16.03$ \\
        PR & $2.92$ & $29.85$ & $30.58$ & $31.14$ & $27.72$ \\
    \bottomrule
    \end{tabular}
    }
    \caption{Performance of removing different tasks. ``QIC'' denotes query intent classification, ``QR'' denotes query reformulation, ``FV'' denotes fact verification, ``Summ.'' denotes summarization, ``CP'' denotes citation prediction, and ``PR'' denotes passage ranking.}
    \label{tab:remove_task}
\end{table}

\begin{table}[t]
    \centering
    \small
    \setlength{\tabcolsep}{4mm}{
    \begin{tabular}{lccccccc}
    \toprule
        Dataset & No FT & \ourdata{} & \textit{w/o} Ds \\
    \midrule
        QReCC (RM) & $14.31$ & $80.65$ & $75.02$ \\
        CANARD & $7.83$ & $83.42$ & $83.33$ \\
        XSum (RM) & $10.39$ & $28.66$ & $12.31$ \\
        MultiNews & $6.70$ & $11.20$ & $12.17$ \\
        Quora (RM) & $4.06$ & $84.26$ & $77.68$ \\
        MS MARCO & $2.92$ & $29.85$ & $29.34$ \\
    \bottomrule
    \end{tabular}
    }
    \caption{Performance of removing several datasets, including TREC-Robust, QReCC, MIMICS-Duo, Climate-FEVER, XSum, Quora, and NQ. Models trained with the ablated dataset is denoted as ``\textit{w/o} Ds''. ``RM'' indicates the dataset is removed from the training set and becomes unseen during test.}
    \label{tab:remove_dataset}
\end{table}
\subsection{Out-of-domain Evaluation}\label{sec:out-of-domain-eval}
Instruction fine-tuned LLMs have demonstrated a remarkable zero-shot performance on unseen tasks~\cite{flan,flanv2}. We also investigate the generalizability of the models after fine-tuned on \ourdata{}. Specifically, we consider the following three scenarios. 

\noindent$\bullet$ {Category-level} generalizability: In this scenario, we exclude an entire category of tasks (\eg, query understanding) from \ourdata{}. Then, we fine-tune the models on the remaining data and test them on all datasets. This experiment can help us understand how distinct categories of tasks relate to each other and contribute to overall model performance.

\noindent$\bullet$ {Task-level} generalizability: In this scenario, we remove specific tasks (\eg, query intent classification) from \ourdata{}. Similarly, we fine-tune the models on the remaining data and evaluate them on all datasets. The goal is to assess whether fine-tuned models can generalize to unseen tasks effectively.

\noindent$\bullet$ {Dataset-level} generalizability: In this scenario, due to the large number of datasets in \ourdata{}, we exclude several datasets at once (including TREC-Robust, QReCC, MIMICS-Duo, Climate-FEVER, XSum, Quora, and NQ) from \ourdata{}. Then, we fine-tune the models on the remaining data and test them on all datasets. This experiment aims to evaluate the fine-tuned models' ability to generalize to unseen datasets within the scope of learned tasks. 

We analyze the experimental result as follows:

(1) In the category-level ablation study (Table~\ref{tab:remove_group}), the models fine-tuned with the full \ourdata{} outperform those trained on ablated versions, verifying the efficacy of comprehensive fine-tuning in improving search task performance. We can also see that models trained on a subset of tasks still surpass the performance of the untrained models. For example, the performance of ``\textit{w/o} Q'' is higher than the untuned model ``(No FT)'' on query understanding tasks. This result indicates that the different task categories are effectively complementary.

(2) Table~\ref{tab:remove_task} shows that the models exhibit task-level generalization. For instance, models fine-tuned without the query intent classification (QIC) task still outperform the untrained ones in this task. This implies that knowledge learned from other search tasks helps understand query intent. Furthermore, the query reformulation (QR) task's performance also drops when the query intent classification task is removed, further supporting the interdependence of these tasks. Overall, task-level generalization indicates that LLMs fine-tuned on \ourdata{} can be better applied to other search tasks.

(3) Some results of the third scenario is illustrated in Table~\ref{tab:remove_dataset} (full results in Appendix~\ref{app:add_result}). Compared to the previous two scenarios, this scenario is much easier for the fine-tuned model as all tasks have been learned during training. Generally, the models exhibit good generalizability among datasets, as evidenced by the superior performance of ``\textit{w/o} Ds'' compared to the untuned model (``No FT'') across all datasets.
We also notice that removing XSum from training leads to improved performance on MultiNews, highlighting the complex relationship between different datasets and suggesting a need for further exploration into the optimal dataset combinations for instruction tuning.

\subsection{Further Analysis}
We also conduct a series of experiments to investigate the impact of different settings in \ourdata{}. All the experiments are conducted based on fine-tuning the LLaMA-2-Chat-7b model.

\begin{figure}[t]
    \centering
    \includegraphics[width=\linewidth]{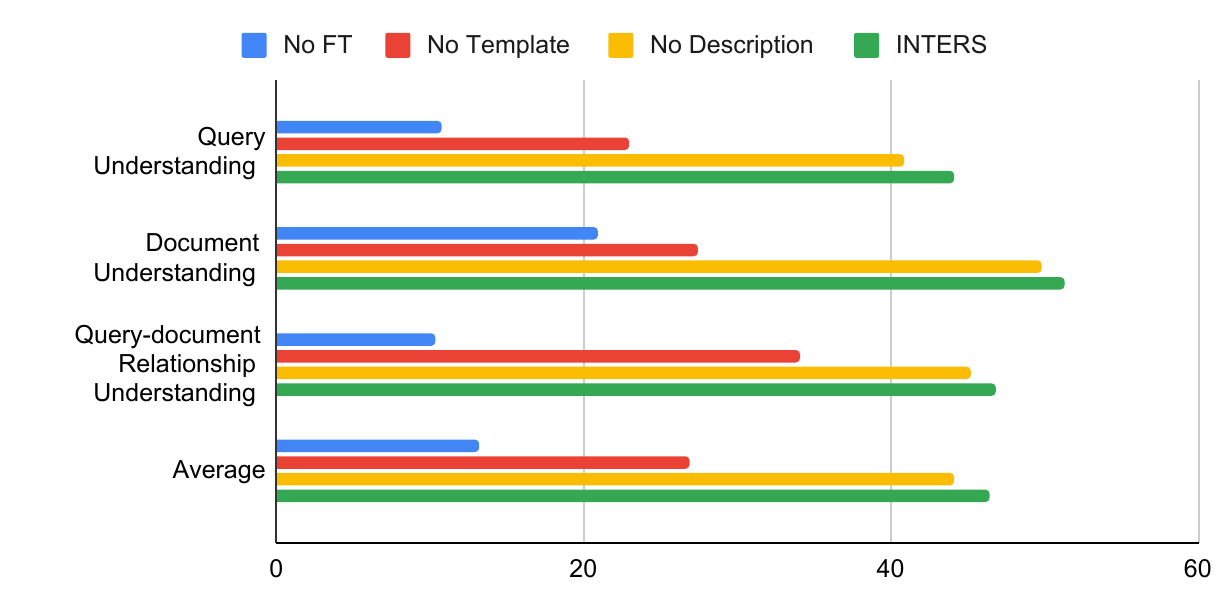}
    \caption{Ablation study result of using no template or no description during training.}
    \label{fig:no-template}
\end{figure}

\begin{table}[t]
    \centering
    \small
    \setlength{\tabcolsep}{1.5mm}{
    \begin{tabular}{lccccc}
    \toprule
    Task & Min & Max & Avg.$\pm$Std. & Random & Unseen \\
    \midrule
        QD & $26.49$ & $29.27$ & $28.55\pm 0.87$ & $29.90$ & $29.20$ \\
        QE & $36.37$ & $40.03$ & $38.10\pm 1.08$ & $37.84$ & $38.33$ \\
        QR & $69.32$ & $70.50$ & $69.93\pm 0.42$ & $69.39$ & $64.58$ \\
        QC & $23.44$ & $24.42$ & $23.97\pm 0.27$ & $23.66$ & $21.38$ \\
        QSG & $10.76$ & $13.08$ & $11.59\pm 0.85$ & $10.76$ & $12.67$\\
        QS & $42.57$ & $55.87$ & $53.48\pm 3.93$ & $50.24$ & $54.78$ \\
        QM & $83.17$ & $86.22$ & $84.62\pm 0.95$ & $85.54$ & $85.92$ \\
        QIC & $47.45$ & $58.30$ & $54.72\pm3.49$ & $53.92$ & $55.18$ \\
        \midrule
        Avg. & $42.61$ & $45.40$ & $44.39\pm 0.84$ & $43.25$ & $44.06$ \\
    \bottomrule
    \end{tabular}
    }
    \caption{Result of using various templates for evaluation. All tasks are from query understanding, and their names are represented by the abbreviations, \eg, ``QD'' denotes query description.}
    \label{tab:templates}
\end{table}

\subsubsection{Impact of Task Description \& Templates}\label{sec:task-description}
\ourdata{} includes a detailed description and 12 distinct templates for each task to enhance task comprehension and increase data diversity. We examine their effectiveness by the following experiments. 

The result shown in Figure~\ref{fig:no-template} demonstrates that the use of task descriptions significantly improves model performance across most datasets. This strongly supports our hypothesis that detailed task descriptions aid in task understanding. Besides, the task description appears to enhance the instruction tuning process, leading to substantial improvements in some cases (\eg, a 51.8\% performance improvement on the MIMICS query clarification dataset as shown in Appendix~\ref{app:add_result}). We speculate that these task descriptions not only clarify individual tasks but also facilitate more effective cross-dataset knowledge transfer.

In the construction of \ourdata{}, a key component is the development of 12 distinct templates for each dataset, aiming at guiding the models in task comprehension. It is also interesting to study the influence of these templates on model performance. At first, we compare the performance when training with or without these templates. 
For the no template setup, we retain the keywords to indicate the different parts of the input. For the example shown in Figure~\ref{fig:construction}, we keep only ``Context: ... Query: ...'' as input. Besides, we follow FLAN and use the \ourdata{} instructions for zero-shot testing (because if we use no template, the model cannot know what task to perform). The results, shown in Figure~\ref{fig:no-template}, reveal that omitting templates leads to suboptimal performance, highlighting the instructional templates' critical role in task learning.

Next, we study the impact of different templates. By default, we use random templates for evaluation, while in this experiment, we use each template to build test samples for query-understanding tasks and compare their performance differences. Besides, to simulate the real application scenario, we manually write a new unseen template for testing. The results are shown in Table~\ref{tab:templates}. We can see that while the model can achieve significantly better performance than the untuned model on any template, template selection is still vital for some tasks, such as query suggestion (maximum $55.87$ vs. minimum $42.57$). This reflects the importance of deliberate template design. Remarkably, models tested on unseen templates can still show superior performance. This demonstrates again that our instruction-tuned models have good robustness and generalizability.

\begin{table}[t]
    \centering
    \small
    \setlength{\tabcolsep}{1.7mm}{
        \begin{tabular}{lccc}
        \toprule
           Task & No FT & FLAN & \ourdata{}-T \\
        \midrule
           Query Intent Classification  & $23.34$ & $24.40$ & $38.55$ \\
           Fact Verification & $48.43$ & $57.67$ & $49.08$ \\
           Citation Prediction & $2.90$ & $4.79$ & $16.03$ \\
        \bottomrule
        \end{tabular}
    }
    \caption{Performance comparison between \ourdata{} and FLAN on three search-related tasks.} 
    \label{tab:vs-flan}
\end{table}

\subsubsection{Comparison with FLAN}\label{sec:flan}
FLAN~\cite{flan,flanv2} is a commonly used dataset for fine-tuning LLMs on NLP tasks. We compare its effectiveness on search-related tasks with that of our \ourdata{}. Given the significantly larger size of FLAN, we randomly sample 200k data examples from it for a fair comparison.\footnote{\url{https://huggingface.co/datasets/Open-Orca/FLAN}} Besides, to ensure fairness, as FLAN does not include the search-related tasks tested in this experiment, we also remove these tasks from \ourdata{} for comparison (denoted as \ourdata{}-T). By this means, both models trained on FLAN and \ourdata{} are evaluated on tasks not seen during training. The results are shown in Table~\ref{tab:vs-flan}. 

We find that both FLAN and \ourdata{} can enhance LLMs' performance on the three tasks, demonstrating again the effectiveness of instruction tuning in unlocking LLM potential for search tasks. Notably, \ourdata{} yields a more substantial improvement in search tasks, particularly in query-document relationship understanding. This is consistent with our expectations, as \ourdata{} is specifically tailored for search tasks. Although the tested tasks are unseen in training, other search-related tasks can provide relevant knowledge for these tasks. Finally, we can see training on FLAN achieves better performance on fact verification. The potential reason is that this task is very close to other NLP tasks included in FLAN, enabling effective knowledge transfer. Unfortunately, we do not observe further improvement by combining FLAN and \ourdata{}, thus the results are omitted.



\subsubsection{Zero-shot vs. Few-shot Demonstrations}\label{sec:few-shot}
\begin{figure}
    \centering
    \begin{subfigure}[b]{.49\linewidth}
         \centering
         \includegraphics[width=\linewidth]{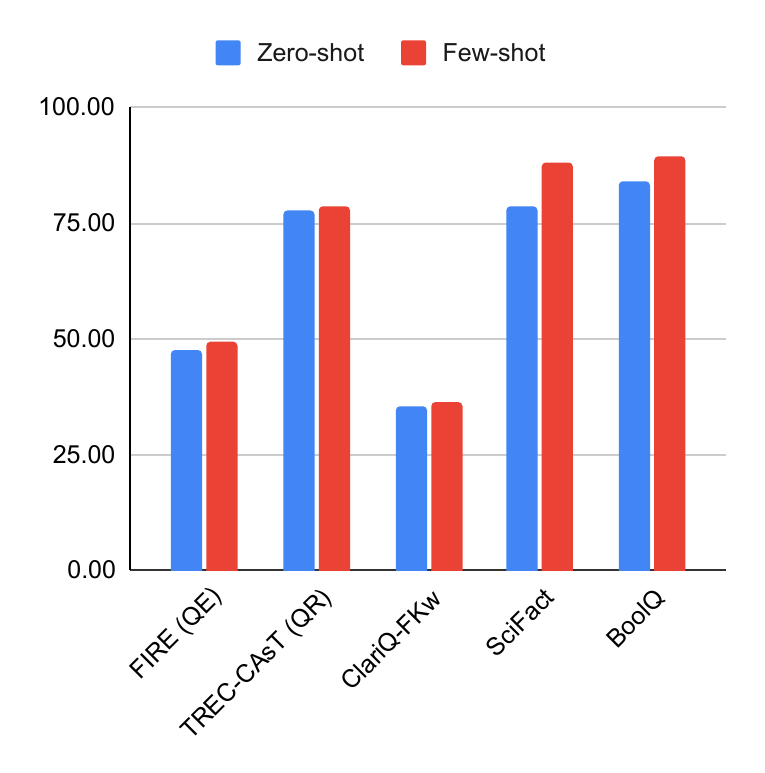}
     \end{subfigure}
     \begin{subfigure}[b]{.49\linewidth}
         \centering
         \includegraphics[width=\linewidth]{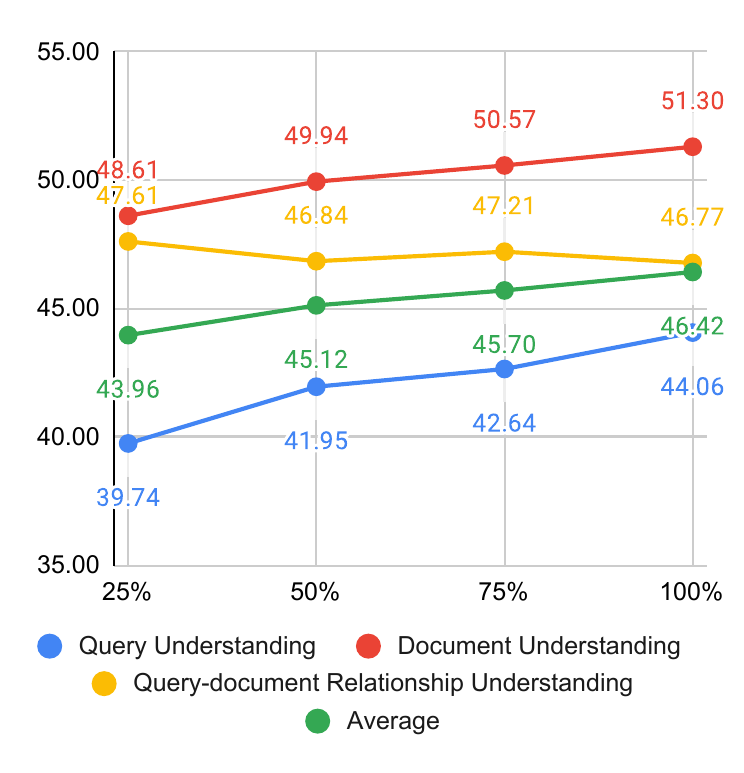}
     \end{subfigure}
    \caption{Performance of using few-shot demonstrations (left) and different data volumes (right).}
    \label{fig:data-volume}
\end{figure}

LLMs have a strong ability of few-shot learning (also known as in-context learning), which enables them to quickly adapt to a wide range of tasks. Given that \ourdata{} comprises a mix of zero-shot and few-shot, it is critical to examine the few-shot performance of the LLMs fine-tuned on \ourdata{}. We choose datasets for few-shot ($n=5$) testing that fit within the models' input length limit (2,048 tokens in our case). The results are shown on the left side of Figure~\ref{fig:data-volume}. Generally, few-shot demonstrations bring a consistent improvement in performance across all datasets, compared to zero-shot scenarios. Few-shot demonstrations are particularly beneficial in tasks with complex output spaces, such as reading comprehension (BoolQ), potentially because these examples help the model better understand the task and output format.

\subsubsection{Impact of Data Volumes}\label{sec:data-volume}
The quantity of training data plays a pivotal role in the success of instruction tuning. To explore this, we conduct experiments using 25\%, 50\%, and 75\% of the data sampled from \ourdata{} for training. The results on the right side of Figure~\ref{fig:data-volume} clearly demonstrate that increasing the volume of instructional data generally enhances model performance. However, the sensitivity to data volume varies across tasks. For instance, while the query understanding task shows consistent performance across data volumes, increasing data volume cannot effectively improve the performance of query-document relationship understanding. 
This highlights the need for further research to optimize the mix and volume of instructional data for diverse tasks.

\subsubsection{Impact of Ranking Strategy}\label{sec:ranking}
In our data construction (Section~\ref{sec:construction}), we consider three typical methods for query-document understanding tasks, namely pointwise, pairwise, and listwise. Consequently, we test fine-tuned models across these different ranking strategies. Due to the limited space, we report the findings directly: the pointwise methods outperform the pairwise, which in turn exceeds the listwise in effectiveness (so we report pointwise performance in our previous experiments). Moreover, models with 7B or fewer parameters cannot handle the listwise evaluation. This may be due to the fact that the listwise method requires comparing multiple documents simultaneously and employs a sliding window method, presenting a complexity beyond the capability of such models. The comprehensive results are presented in Appendix~\ref{app:rerank}.

\section{Conclusion}
In this paper, we investigated the application of instruction tuning to augment the capabilities of LLMs in performing search tasks. Our instruction tuning dataset \ourdata{} demonstrated its effectiveness in consistently enhancing the performance of various open-sourced LLMs across both in-domain and out-of-domain settings. Our extensive experiments delved into several critical aspects, including the structure and design of instructions, the effects of few-shot learning, and the significance of data volumes in instruction tuning. It is our aspiration that this paper will serve as a catalyst for further research in the realm of LLMs, particularly in their application to IR tasks, and will encourage continued exploration into the optimization of instruction-based methods for enhancing the performance of these models.

\section*{Limitation}
In this study, we introduce a novel dataset specifically designed for instruction tuning on search-related tasks, along with models fine-tuned using this dataset. We acknowledge several limitations in our current work that offer avenues for future research. 

First, while our dataset encompasses 20 tasks across 43 datasets, there are still many tasks and datasets that have not been included. Our experimental results suggest that incorporating more data sources can improve the richness and diversity of the dataset, potentially improving overall model performance. Second, due to our limited resources, we cannot conduct experiments with larger LLMs, such as those with 13B, 30B, or even 70B parameters. 
It is interesting to investigate the influence of instruction tuning on these models and compare their search performance with close-sourced LLMs such as GPT-4 if possible. Third, in the query-document relationship understanding part, we only consider the reranking architecture. It is valuable to explore the application of LLMs to other architectures, such as retrieval. 

\section*{Acknowledgments}
This work was supported by Beijing Natural Science Foundation No. L233008, National Natural Science Foundation of China No. 62272467,  the fund for building world-class universities (disciplines) of Renmin University of China, and Public Computing Cloud, Renmin University of China. The work was partially done at the Engineering Research Center of Next-Generation Intelligent Search and Recommendation, MOE.

\balance

\appendix
\section{Details about Tasks \& Datasets}\label{app:task}
We introduce the tasks and datasets as follows. Some examples of our generated data are shown in Appendix~\ref{app:example}.
\subsection{Query Understanding}
In IR, a query is a user-initiated request for information, typically composed of keywords, phrases, or natural language questions. It aims at retrieving relevant information from a retrieval system (\eg, a search engine). The effectiveness of a query is measured by its ability to accurately reflect the user's intent and retrieve the most relevant documents. During the retrieval process, query understanding is a critical component in determining the efficiency and user satisfaction of the IR systems. Therefore, we collect a group of tasks addressing aspects of query understanding to enhance LLMs' capability of understanding the semantics of queries and capturing the underlying user search intent. Specifically, we consider the following eight tasks.

\noindent$\bullet$ \textbf{Query description}: The query description task involves describing the documents potentially relevant to a user-provided query. Queries typically comprise keywords reflecting the user's information needs. The objective of the task is to articulate the characteristics and content of documents that would be considered pertinent to these keywords, aiding in the understanding and retrieval of relevant information. We use the following four datasets: GOV2,\footnote{\url{https://ir-datasets.com/gov2.html\#gov2}} TREC-Robust~\cite{robust04,robust05}, TREC-COVID~\cite{TREC-COVID}, and FIRE 08, 10-12.\footnote{\url{https://www.isical.ac.in/~fire/data.html}} Taking the dataset GOV2 as an example, the query and its description are directly provided.

\noindent$\bullet$ \textbf{Query expansion}: The query expansion task involves elaborating an original, brief query into a longer, more detailed version while preserving the original search intent. This process enhances the search engine's understanding of the user's needs, leading to more accurate and relevant document retrieval. We use the following seven datasets: GOV2, TREC-Robust, TREC-COVID, FIRE, Query2Doc~\cite{wang2023query2doc}, TREC-CAsT~\cite{TREC-CAsT}, and TREC-Web 09-14.\footnote{\url{https://trec.nist.gov/data/webmain.html}} Taking the dataset GOV2 as an example, the query and its expansion are directly provided.

\noindent$\bullet$ \textbf{Query reformulation}: The query reformulation task enhances user-input queries to be more explicit and comprehensible for search engines. It addresses omissions typical of user queries, which often exclude common sense or contextually implied information. The refined query, therefore, includes all the necessary details to guide the search engine towards retrieving the most relevant documents. We use the following datasets: CODEC~\cite{CODEC}, QReCC~\cite{QReCC}, CANARD~\cite{CANARD}, TREC-CAsT, and GECOR~\cite{GECOR}. Taking the dataset CODEC as an example, the queries and their reformulations are provided in two files, and we can connect them by query IDs.

\noindent$\bullet$ \textbf{Query intent classification}: User queries can have various search intents, such as informational (seeking knowledge about a topic), transactional (aiming to purchase a product), or navigational (looking to find a specific website). The intents can also be more specific in certain scenarios. Accurately discerning the type of intent behind a query is crucial for search engines to tailor and refine their results effectively. We use the following three datasets: ORCAS-I~\cite{ORCAS-I}, MANtIS~\cite{MANtIS}, and TREC-Web 09-14. Taking the dataset ORCAS-I as an example, each query is associated with an attribute ``query type'', which indicates the query's intent.

\noindent$\bullet$ \textbf{Query clarification}: The query clarification task addresses unclear or ambiguous user queries by asking for further details or providing clarification options. This process helps refine the query, resulting in clearer and more precise search terms for improved search engine results. We use the following datasets: MIMICS~\cite{MIMICS}, MIMICS-Duo~\cite{MIMICS-Duo}, ClariQ-FKw~\cite{ClariQ-FKw}, and RaoCQ~\cite{RaoCQ}. Taking the dataset MIMICS as an example, each query is labeled with a list of clarification options.

\noindent$\bullet$ \textbf{Query matching}: The query matching task involves determining whether two queries or texts, despite differing in expression, convey the same meaning. This is crucial in search tasks where identifying synonymous queries can enhance the relevance and accuracy of results. We use the dataset: MSRP.\footnote{\url{https://www.microsoft.com/en-us/download/details.aspx?id=52398}} It provides a label for each pair of sentences, indicating whether they convey identical content.

\noindent$\bullet$ \textbf{Query subtopic generation}: The query subtopic generation task addresses the ambiguity of web searches by identifying and presenting various aspects of the initial query. This approach aids search engines in understanding the query's breadth, leading to more diverse and relevant search results. We use the dataset: TREC-Web 09-14. It contains subtopic annotations for queries. 

\noindent$\bullet$ \textbf{Query suggestion}: In search sessions, users often input a series of queries to fulfill a specific information need. The query suggestion task aims to analyze these queries and associated search behaviors to understand the user's intent and predict the next likely query, thereby enhancing the search experience. We use the AOL dataset.\footnote{The AOL dataset has been officially withdrawn. However, as it is the most commonly used dataset for query suggestion, we still include it in \ourdata{}.} 
This dataset contains a large number of search sessions. Within each session, a query at a specific position is randomly chosen to represent the ``next query''. Subsequently, the preceding queries in the session, optionally inclusive of the clicked documents, are utilized as the search context. 

\subsection{Document Understanding}
In IR, a document refers to any piece of information that can be retrieved in response to a query, such as web pages in search engines. Document understanding is the process by which an IR system interprets and comprehends the content and context of these documents. The importance of document understanding lies in its direct impact on the effectiveness and accuracy of information retrieval. Enhanced document understanding leads to better search results, more effective organization of information, and an overall more efficient and user-friendly retrieval process. Therefore, we collect the following four tasks to enhance LLMs' capability of document understanding.

\noindent$\bullet$ \textbf{Fact verification}: The fact verification task involves assessing whether a claim is supported or refuted by the given evidence. It requires a clear analysis of the relationship between the claim and the evidence, with a careful check to determine if there is sufficient information for a conclusive judgment. Such detailed understanding aids search engines in achieving a deeper comprehension of the documents, enhancing their ability to deliver accurate and relevant results. We use the three datasets: FEVER~\cite{FEVER}, Climate-FEVER~\cite{Climate-FEVER}, and SciFact~\cite{SciFact}. Taking the dataset FEVER as an example, it provides claims, their labels, and the corresponding evidences.

\noindent$\bullet$ \textbf{Summarization}: The text summarization task seeks to create a concise summary of one or more lengthy documents, encapsulating all vital information while omitting extraneous details. The summary must accurately reflect the content of the original documents without introducing any new information. Achieving this necessitates a profound understanding of the documents, which can significantly enhance the performance of search engines by providing distilled, relevant content. We use four datasets: CNN/DM~\cite{CNN/DM}, WikiSum~\cite{WikiSum}, Multi-News~\cite{Multi-News}, and XSum~\cite{XSum}. Taking the dataset CNN/DM as an example, it provides articles and their summaries.

\noindent$\bullet$ \textbf{Reading comprehension}: The reading comprehension task requires generating an answer to a question using information from a given context. It necessitates a deep understanding of the text's context and semantics, enabling search engines to more accurately rank the relevance of retrieved documents based on this nuanced comprehension. We use the following six datasets: SQuAD~\cite{SQUAD-1.0}, HotpotQA~\cite{HotpotQA}, MS MARCO~\cite{MS-MARCO}, TriviaQA~\cite{TriviaQA}, BoolQ~\cite{BoolQ}, and WebGLM-QA~\cite{webglm}. Taking the dataset SQuAD as an example, it provides questions, their answers, and the corresponding context.
 
\noindent$\bullet$ \textbf{Conversational question-answering}: Conversational question-answering involves responding to a series of interrelated questions based on a given context. As these questions might build upon shared information, some details may be implicitly understood rather than explicitly stated. By comprehensively understanding and analyzing this dialogue structure, search engines can enhance their interpretation of user queries and their connections to relevant documents, thereby improving result accuracy and relevance. We use these two datasets: CoQA~\cite{CoQA} and QuAC~\cite{CoQA}. Taking the dataset CoQA as an example, each data sample contains a story, a series of questions about the story, and the corresponding answers.

\subsection{Query-document Relationship Understanding}
Query-document relationship understanding in information retrieval is the process of determining how well the content of a document matches or satisfies the intent behind a user's query. This involves interpreting the query's semantics, context, and purpose, and then assessing the relevance of documents based on how closely they correspond to these aspects. It is the core task of information retrieval. The relationship between queries and documents varies in different scenarios. For example, in question answering, the model needs to understand the relationship between the question and its potential relevant materials. In fact checking, the model is required to examine the relationship between the claim and its supporting evidence. 

We use the MS MARCO passage ranking dataset and the datasets in the BEIR~\cite{BEIR} benchmark across multiple domains (such as bio-medical, finance, and social media), which includes Touché-2020~\cite{Touche-2020}, ArguAna~\cite{ArguAna}, TREC-COVID, NFCorpus~\cite{NFCorpus}, SciDocs~\cite{SciDocs}, Quora,\footnote{\url{https://quoradata.quora.com/First-Quora-Dataset-Release-Question-Pairs}} CQADupStack~\cite{CQADupStack}, DBPedia~\cite{DBPedia}, FEVER, Climate-FEVER, SciFact~\cite{SciFact}, NQ~\cite{NQ}, FiQA~\cite{FiQA}, and HotpotQA. Note that some datasets do not have queries in the training set, so we use the generated queries provided by BEIR.\footnote{\url{https://huggingface.co/BeIR}}

It is important to recognize the variety of architectures available for modeling the query-document relationship. In this research, we focus on the reranking architecture, which is the most straightforward way to apply LLMs. More details about applying LLMs to document reranking are provided in Appendix~\ref{app:rerank}. The primary objective of document reranking is to rerank a list of candidate documents according to their relevance to the user's query. The most relevant documents, those that best cover the user's information needs, are ranked at the top of the list. In our experiments, we use the documents retrieved by BM25~\cite{bm25} as the candidates.

The \textbf{statistics} of all datasets and their \textbf{evaluation metrics} are reported in Table~\ref{tab:statistics}.

\subsection{Licenses}\label{app:license}
We plan to release our data under the license of CC BY-SA 4.0.\footnote{\url{https://creativecommons.org/licenses/by-sa/4.0/}} The authors of 10 out of the 43 datasets in \ourdata{} (FIRE, TREC-Web, MANtIS, ClariQ-FKw, RaoCQ, AOL, Climate-FEVER, WikiSum, TriviaQA, and WebGLM-QA) do not report the dataset license in the paper or a repository. The rest is over view as follows:

$\bullet$ Apache License 2.0 license: GOV2, TREC-Robust, CODEC, CNN/DM

$\bullet$ MIT license: TREC-CAsT, GECOR, ORCAS-I, MIMICS, MIMICS-Duo, XSum

$\bullet$ CC BY 4.0: Query2Doc, MSRP, SQuAD

$\bullet$ CC BY-SA 4.0: CANARD, BEIR, HotpotQA, QuAC

$\bullet$ CC BY-SA 3.0: QReCC, FEVER, BoolQ

$\bullet$ CC BY-NC 2.0: SciFact

$\bullet$ Provided under the ``Dataset License Agreement'': TREC-COVID, Multi-News, MS MARCO

Note that CoQA contains several datasets under different licenses. They are listed on the HuggingFace page.\footnote{\url{https://huggingface.co/datasets/stanfordnlp/coqa}}

\section{In-domain Evaluation Details}\label{app:eval}
In this evaluation, we split the full dataset into training, validation, and test sets. The split process is designed based on the size and structure of the original datasets.
Specifically, if the original datasets do not contain a test set, then: 
For original datasets with over 10,400 samples, we randomly select 10,000 samples for constructing training data, 200 samples for validation, and 200 samples for testing. In instances where the datasets comprise between 2,000 and 10,400 samples, we randomly select 200 samples each for validation and testing, with the remainder constructing the training samples. For smaller datasets contain fewer than 2,000 samples, we use the ratio of 8:1:1 to obtain the training, validation, and test sets. When the original dataset includes a test set, we use the test set to construct test samples, extracting only the validation set from the samples in the training dataset. The extraction rule is similar to the previous case.

\begin{figure}[t]
    \centering
    \includegraphics[width=\linewidth]{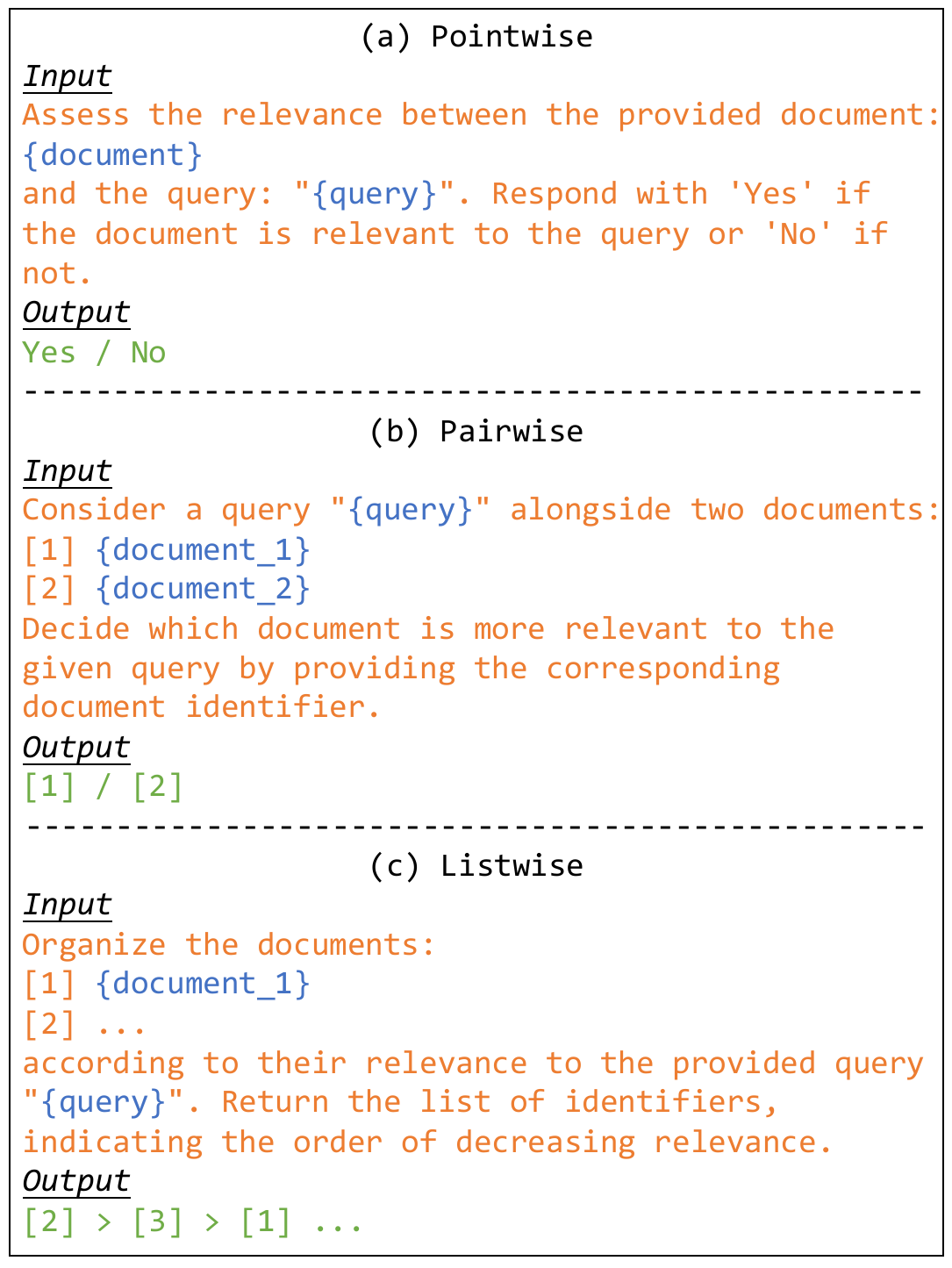}
    \caption{Three typical methods of applying LLMs for reranking.}
    \label{fig:ranking_llm}
\end{figure}

\begin{table}[t]
    \centering
    \tiny
    \setlength{\tabcolsep}{1.3mm}{
    \begin{tabular}{lllccc}
    \toprule
    Dataset & Model & Metric & Pointwise & Pairwise & Listwise \\
    \midrule
    Touché-2020 & LLaMA & MRR@10 & 0.0667 & \textbf{0.2010} & 0.1930 \\
    Touché-2020 & LLaMA & NDCG@10 & 0.0082 & 0.0645 & \textbf{0.0663} \\
    Touché-2020 & INTERS-LLaMA & MRR@10 & 0.2037 & \textbf{0.3173} & 0.2186 \\
    Touché-2020 & INTERS-LLaMA & NDCG@10 & 0.1132 & \textbf{0.1196} & 0.0802 \\
    TREC-COVID & LLaMA & MRR@10 & 0.6209 & \textbf{0.6842} & 0.6662 \\
    TREC-COVID & LLaMA & NDCG@10 & 0.3849 & 0.4103 & \textbf{0.4162} \\
    TREC-COVID & INTERS-LLaMA & MRR@10 & 0.8317 & \textbf{0.8609} & 0.7048 \\
    TREC-COVID & INTERS-LLaMA & NDCG@10 & \textbf{0.6337} & 0.5874 & 0.4208 \\
    NFCorpus & LLaMA & MRR@10 & \textbf{0.3474} & 0.1646 & 0.1806 \\
    NFCorpus & LLaMA & NDCG@10 & \textbf{0.1956} & 0.0703 & 0.0715\\
    NFCorpus & INTERS-LLaMA & MRR@10 & \textbf{0.3766} & 0.2787 & 0.3038 \\
    NFCorpus & INTERS-LLaMA & NDCG@10 & \textbf{0.2272 }& 0.1120 & 0.1191 \\
    DBPedia & LLaMA & MRR@10 & 0.1777 & \textbf{0.2096} & 0.2091 \\
    DBPedia & LLaMA & NDCG@10 & 0.0637 & \textbf{0.0703} & 0.0687 \\
    DBPedia & INTERS-LLaMA & MRR@10 & \textbf{0.7386} & 0.4877 & 0.3009 \\
    DBPedia & INTERS-LLaMA & NDCG@10 & \textbf{0.4026} & 0.2004 & 0.1138 \\
    SciFact & LLaMA & MRR@10 & 0.0133 & \textbf{0.0203} & 0.0179 \\
    SciFact & LLaMA & NDCG@10 & 0.0224 & \textbf{0.0337} & 0.0300 \\
    SciFact & INTERS-LLaMA & MRR@10 & \textbf{0.7307} & 0.2357 & 0.2327 \\
    SciFact & INTERS-LLaMA & NDCG@10 & \textbf{0.7522} & 0.2409 & 0.2399 \\
    FiQA & LLaMA & MRR@10 & \textbf{0.0437} & 0.0317 & 0.0317 \\
    FiQA & LLaMA & NDCG@10 & \textbf{0.0355} & 0.0274 & 0.0274 \\
    FiQA & INTERS-LLaMA & MRR@10 & \textbf{0.4437} & 0.1803 & 0.0700 \\
    FiQA & INTERS-LLaMA & NDCG@10 & \textbf{0.3649} & 0.1266 & 0.0514 \\
    \midrule
    Average & LLaMA & MRR@10 & \textbf{0.2247} & 0.2186 & 0.2164 \\
    Average & LLaMA & NDCG@10 & \textbf{0.1282} & 0.1127 & 0.1134 \\
    Average & INTERS-LLaMA & MRR@10 & \textbf{0.5542} & 0.3934 & 0.3051 \\
    Average & INTERS-LLaMA & NDCG@10 & \textbf{0.4156} & 0.2312 & 0.1709 \\
    \bottomrule
    \end{tabular}
    }
    \caption{Performance of using different reranking methods on several query-document relationship understanding tasks.}
    \label{tab:reranking_methods}
\end{table}

\section{LLMs for Reranking}\label{app:rerank}
To apply LLMs for the document reranking task, there are three typical methods: pointwise, pairwise, and listwise (shown in Figure~\ref{fig:ranking_llm}). They use different prompts to perform reranking. A brief overview of these methods is presented below, with further details accessible in the literature~\cite{llm4ir, DBLP:journals/corr/abs-2306-17563}.

(1) \textbf{Pointwise methods} measure the relevance between a query and a single document. As illustrated in Figure~\ref{fig:ranking_llm} (a), a common method is prompting the LLMs to judge whether a query and a document are relevant. The relevance score is computed based on the generation probability of ``Yes'' and ``No'' tokens: $r=p_{\rm yes}/(p_{\rm yes} + p_{\rm no})$.

(2) \textbf{Pairwise methods} require LLMs to determine which of two documents is more relevant to the given query, as shown in Figure~\ref{fig:ranking_llm} (b). To get a ranking list of all candidate documents, aggregation methods, such as PRP-Allpair~\cite{DBLP:journals/corr/abs-2306-17563}, are applied.

(3) \textbf{Listwise methods} directly prompt LLMs to generate a reranked list of documents, as shown in Figure~\ref{fig:ranking_llm} (c). However, due to the limited input length of LLMs, it is often impractical to include all candidate documents in a single prompt. To address this, a sliding window strategy is commonly applied~\cite{rankgpt}.

To support various application scenarios, we consider all three methods when collecting the templates for the query-document relationship understanding tasks in \ourdata{}. Concretely, we collect four distinct templates for each of these methods. The performance comparison of different methods has been presented in Table~\ref{tab:reranking_methods}. Note that we only report the results on a select number of datasets due to the substantial computational costs of pairwise methods. From the results, we can see:

First, no matter which reranking method is used, \ourdata{} can consistently improve LLMs' performance on query-document relationship understanding tasks, reflecting its broad applicability. Second, comparing the three methods, the pointwise methods generally outperform the pairwise methods, which in turn exceed the listwise methods in effectiveness. Moreover, models with 7B or fewer parameters cannot handle the listwise evaluation. This may be due to the fact that the listwise method requires comparing multiple documents simultaneously and employs a sliding window method, presenting a complexity beyond the capability of such models. Third, in terms of inference cost, the pairwise method is the most resource-intensive. This is attributed to its requirement for pairwise document comparisons and an additional algorithm for deriving the final result. The cost of listwise methods relies on the sliding window algorithm, but its performance relies on the quality of the initial ranking list~\cite{rankgpt}. Based on these observations, we consider that the pointwise method is the most suitable one for query-document relationship understanding tasks on LLMs with 7B or fewer parameters, which provides a good balance between efficacy and computational costs.

\section{Backbone Models \& Implementation Details}\label{app:implementation}
We employ four LLMs in different sizes, ranging from 1B parameters to 7B parameters:

\noindent$\bullet$ \textbf{Falcon-RW-1B}~\cite{falcon-rw-1b} is a language model developed by the Technology Innovation Institute, trained on 600B tokens of English data. 
The model is designed for researching large language models and the impact of adequately filtered and deduplicated web data on their properties, such as fairness, safety, limitations, and capabilities.\footnote{\url{https://huggingface.co/tiiuae/falcon-rw-1b}} 

\noindent$\bullet$ \textbf{Minima-2-3B}~\cite{minima-2-3b} is a novel language model designed to achieve a new compute-performance frontier on common benchmarks by distilling knowledge from a large teacher language model (LLaMA-2-7B). The model uses a data mixture of 126 billion tokens from various sources for distillation.\footnote{\url{https://huggingface.co/GeneZC/MiniMA-2-3B}}

\noindent$\bullet$ \textbf{Mistral-7B}~\cite{mistral-7b} is a language model engineered for superior performance and efficiency. It leverages mechanisms such as grouped-query attention~\cite{gqa} and sliding window attention~\cite{swa1, swa2} to outperform other language models in various benchmarks.\footnote{\url{https://huggingface.co/mistralai/Mistral-7B-v0.1}}

\noindent$\bullet$ \textbf{LLaMA-2-7B}~\cite{llama2} is language model trained on around 2T tokens. It has shown exceptional performance across multiple benchmark tests and has been widely used for LLM research. In our experiments, we find that the LLaMA-2-Chat model performs slightly better than the LLaMA-2-Base after fine-tuning (the result is reported in Section~\ref{sec:in-domain-eval}). Therefore, we use LLaMA-2-Chat in our main experiments and further investigation.\footnote{\url{https://huggingface.co/meta-llama/Llama-2-7b}, \url{https://huggingface.co/meta-llama/Llama-2-7b-chat-hf}}

For all backbone models, we used their publicly available checkpoints on Huggingface. The fine-tuning process was implemented using PyTorch and Colossal-AI frameworks~\cite{colossal-ai}. To optimize memory usage and accelerate training, we applied Deepspeed ZeRO stage 2~\cite{deepspeed} and BFloat16 mixed precision techniques. Additionally, Flash attention~\cite{flash-attention} was used to further improve training efficiency. The training was conducted with a batch size of 32, a learning rate of 1e-5, and a maximum length setting of 2,048 tokens. Though some backbone models support longer inputs, we limit the input length to reduce training costs. All models were trained on 8 Tesla A100-40G GPUs. It is important to note that the hyperparameters were set based on empirical observations, as the primary aim was to validate the effectiveness of \ourdata{}. Comprehensive hyperparameter tuning was beyond the scope of this study due to resource limitations.

\begin{table}[t]
    \centering
    \small
    \begin{tabular}{lcccc}
    \toprule
    Task & LLaMA-2-7B-Chat & 70B & GPT-4 \\
    \midrule
        QD & 10.21 & 12.65 & 9.01 \\
        QE & 9.29 & 9.78 & 17.38 \\
        QR & 7.82 & 8.99 & 25.24 \\
        QC & 2.73 & 3.30 & 4.49 \\
        QSG & 0 & 0 & 0 \\
        QS & 3.83 & 2.67 & 2.91 \\
        CQA & 0.42 & 1.33 & 1.48 \\
        Summ & 11.11 & 14.41 & 18.81 \\
        RC & 13.01 & 25.29 & 32.40 \\
    \bottomrule
    \end{tabular}
    \caption{Results of larger models on query understanding and document understanding tasks.}
    \label{tab:larger}
\end{table}

\section{Comparison with Larger Models}\label{app:larger}
We also attempt to evaluate the zero-shot performance of GPT-4 and LLaMA-2-70B. However, due to the limited computational resources, we only evaluate them on query understanding and document understanding tasks. Besides, as some tasks require generation logits (such as BoolQ) for computing evaluation metrics, we do not include them in this evaluation. The results are shown in Table~\ref{tab:larger}. From the results, we can observe that while larger LMs generally perform better on NLP-relevant tasks (such as summarization), they still struggle with IR tasks (such as query clarification). This highlights again the importance of our INTERS dataset for IR tasks.

\begin{table*}[t]
    \centering
    \tiny
    \begin{tabular}{lllrrrrrr}
    \toprule
        \textbf{Task} & \textbf{Dataset} & \textbf{Metrics} & \textbf{\# Examples} & \textbf{Avg \#In} & \textbf{Avg \#Out} \\
    \midrule
        Query Description & GOV2 & BLEU-1\&2, \underline{ROUGE-L} & 900 & 308.07 & 57.90 \\
        Query Description & TREC-Robust & BLEU-1\&2, \underline{ROUGE-L} & 1,794 & 280.38 & 48.18 \\
        Query Description & TREC-COVID & BLEU-1\&2, \underline{ROUGE-L} & 300 & 258.74 & 33.13 \\
        Query Description & FIRE & BLEU-1\&2, \underline{ROUGE-L} & 1,200 & 290.38 & 46.58 \\
        Query Expansion & GOV2 & BLEU-1\&2, \underline{ROUGE-L} & 900 & 168.71 & 15.77 \\
        Query Expansion & TREC-Robust & BLEU-1\&2, \underline{ROUGE-L} & 1,800 & 189.72 & 20.54 \\
        Query Expansion & TREC-COVID & BLEU-1\&2, \underline{ROUGE-L} & 300 & 193.50 & 17.59 \\
        Query Expansion & FIRE & BLEU-1\&2, \underline{ROUGE-L} & 1,200 & 197.39 & 18.83 \\
        Query Expansion & Query2Doc & BLEU-1\&2, \underline{ROUGE-L} & 62,400 & 378.88 & 81.21 \\
        Query Expansion & Trec-CAsT & BLEU-1\&2, \underline{ROUGE-L} & 300 & 182.64 & 17.39 \\
        Query Expansion & TREC-Web & BLEU-1\&2, \underline{ROUGE-L} & 1,506 & 163.57 & 12.50 \\
        Query Reformulation & CODEC & Precision, Recall, \underline{F1} & 236 & 853.89 & 74.29 \\
        Query Reformulation & QReCC & BLEU-1\&2, \underline{ROUGE-L} & 62,395 & 644.02 & 15.66 \\
        Query Reformulation & CANARD & BLEU-1\&2, \underline{ROUGE-L} & 30,437 & 666.32 & 16.43 \\
        Query Reformulation & TREC-CAsT & BLEU-1\&2, \underline{ROUGE-L} & 606 & 444.37 & 14.40 \\
        Query Reformulation & GECOR & BLEU-1\&2, \underline{ROUGE-L} & 4,056 & 559.53 & 12.27 \\
        Query Clarification & MIMICS & EM-Precision, Recall, \underline{F1} & 16,734 & 153.83 & 21.06 \\
        Query Clarification & MIMICS-Duo & EM-Precision, Recall, \underline{F1} & 5,484 & 172.27 & 22.43 \\
        Query Clarification & ClariQ-FKw & BLEU-1\&2, \underline{ROUGE-L} & 13,086 & 142.50 & 12.47 \\
        Query Clarification & RaoCQ & BLEU-1\&2, \underline{ROUGE-L} & 2,759 & 854.22 & 15.33 \\
        Query Subtopic Generation & TREC-Web & Precision, Recall, \underline{F1} & 1,506 & 321.30 & 74.82 \\
        Query Suggestion & AOL & BLEU-1\&2, \underline{ROUGE-L} & 62,400 & 202.07 & 5.18 \\
        Query Matching & MSRP & Accuracy, \underline{F1} & 25,656 & 325.13 & 2.00 \\
        Query Intent Classification & MANtIS & \underline{Precision@1} & 6,062 & 1,109.86 & 3.81 \\
        Query Intent Classification & ORCAS-I & Accuracy, \underline{F1} & 6,000 & 242.26 & 3.36 \\
        Query Intent Classification & TREC-Web & Accuracy, \underline{F1} & 1,200 & 224.34 & 3.66 \\
        \midrule
        \midrule
        Fact Verification & FEVER & Accuracy, \underline{F1} & 61,932 & 547.03 & 2.29 \\
        Fact Verification & Climate-FEVER & Accuracy, \underline{F1} & 8,544 & 1,133.29 & 2.88 \\
        Fact Verification & SciFact & Accuracy, \underline{F1} & 4,638 & 618.58 & 2.34 \\
        Conversational QA & CoQA & \underline{Exact Match} & 19,741 & 1,208.52 & 80.81 \\
        Conversational QA & QuAC & \underline{Exact Match} & 19,874 & 1,267.01 & 124.72 \\
        Summarization & CNN/DM & ROUGE-1\&2, \underline{ROUGE-L} & 21,883 & 823.92 & 301.19 \\
        Summarization & XSum & ROUGE-1\&2, \underline{ROUGE-L} & 31,510 & 1,057.63 & 135.22 \\
        Summarization & WikiSum & ROUGE-1\&2, \underline{ROUGE-L} & 6,874 & 2,101.13 & 422.31 \\
        Summarization & Multi-News & ROUGE-1\&2, \underline{ROUGE-L} & 5,339 & 3,106.26 & 285.37 \\
        Reading Comprehension & SQuAD & \underline{F1} & 62,336 & 858.54 & 5.74 \\
        Reading Comprehension & HotpotQA & \underline{F1} & 62,400 & 595.62 & 5.46 \\
        Reading Comprehension & MS MARCO & \underline{F1} & 40,029 & 1,314.41 & 24.82 \\
        Reading Comprehension & BoolQ & Accuracy, \underline{F1} & 62,384 & 652.50 & 2.00 \\
        Reading Comprehension & WebGLM-QA & BLEU-1\&2, \underline{ROUGE-L} & 29,164 & 1,107.86 & 140.69 \\
        Reading Comprehension & Trivia-QA & \underline{F1} & 34,140 & 1,312.96 & 9.32 \\
        \midrule
        \midrule
        General Retrieval & MS MARCO & MRR@10, \underline{NDCG@10} & 65,909 & 816.71 & 4.25 \\
        Argument Retrieval & Touché-2020 & MRR@10, \underline{NDCG@10} & 21,951 & 992.36 & 4.46 \\
        Argument Retrieval & ArguAna & MRR@10, \underline{NDCG@10} & 42,736 & 1,077.62 & 4.06 \\
        Biomedical Retrieval & TREC-COVID & MRR@10, \underline{NDCG@10} & 31,476 & 1,127.98 & 4.38 \\
        Biomedical Retrieval & NFCorpus & MRR@10, \underline{NDCG@10} & 4,508 & 1,185.16 & 3.79 \\
        Article Retrieval & SciDocs & MRR@10, \underline{NDCG@10} & 41,043 & 1,090.32 & 3.82 \\
        Duplicate Question Retrieval & Quora & MRR@10, \underline{NDCG@10} & 43,930 & 589.70 & 7.20 \\
        Duplicate Question Retrieval & CQADupStack & MRR@10, \underline{NDCG@10} & 88,934 & 1,117.72 & 4.43 \\
        Entity Retrieval & DBPedia & MRR@10, \underline{NDCG@10} & 470 & 909.46 & 3.59 \\
        Fact Retrieval & FEVER & MRR@10, \underline{NDCG@10} & 35,201 & 1,131.90 & 5.20 \\
        Fact Retrieval & Climate-FEVER & MRR@10, \underline{NDCG@10} & 57,672 & 945.14 & 4.11 \\
        Fact Retrieval & SciFact & MRR@10, \underline{NDCG@10} & 1,963 & 1,179.06 & 7.70 \\
        Supporting Evidence Retrieval & NQ & MRR@10, \underline{NDCG@10} & 43,963 & 944.69 & 5.33 \\
        Supporting Evidence Retrieval & FiQA & MRR@10, \underline{NDCG@10} & 20,988 & 1,063.95 & 5.73 \\
        Supporting Evidence Retrieval & Hotpot-QA & MRR@10, \underline{NDCG@10} & 63,441 & 934.56 & 7.41 \\
    \bottomrule
    \end{tabular}
    \caption{The statistics of all datasets. ``\#In'' and ``\#Out'' represent the number of tokens in the input and output with the LLaMA's tokenizer. The underlined metric is used in the figures of the main paper.}
    \label{tab:statistics}
\end{table*}

\section{Additional Results}\label{app:add_result}
We present the full evaluation results in Table~\ref{tab:query_result} -- Table~\ref{tab:qd_result2}.

\begin{table*}[t]
    \centering
    \tiny
    \setlength{\tabcolsep}{1.3mm}{
    \begin{tabular}{llcccccccccccccc}
    \toprule
    &  & \multicolumn{2}{c}{LLaMA-2-Base} & \multicolumn{6}{c}{LLaMA-2-Chat} & \multicolumn{2}{c}{Falcon} & \multicolumn{2}{c}{Minima} & \multicolumn{2}{c}{Mistral} \\
    \cmidrule(lr){3-4}\cmidrule(lr){5-10}\cmidrule(lr){11-12}\cmidrule(lr){13-14}\cmidrule(lr){15-16}
    Task \& Dataset & Metric & Vanilla & +\ourdata{} & Vanilla & +\ourdata{} & +25\% & +50\% & +75\% & +FLAN & Vanilla & +\ourdata{} & Vanilla & +\ourdata{} & Vanilla & +\ourdata{} \\
    \midrule
    \multicolumn{16}{l}{\textbf{{Query Description}}} \\
    GOV2 
     & BLEU-1 & 0.0687 & 0.2143 & 0.1293 & 0.1884 & 0.2172 & 0.2324 & 0.2008 & 0.1057 & 0.0074 & 0.1847 & 0.0308 & 0.1761 & 0.0559 & 0.1894 \\
     & BLEU-2 & 0.0285 & 0.1002 & 0.0611 & 0.0971 & 0.1030 & 0.1096 & 0.0956 & 0.0495 & 0.0021 & 0.0793 & 0.0074 & 0.0707 & 0.0229 & 0.0933 \\
     & ROUGE-L & 0.1197 & 0.2057 & 0.1333 & 0.2063 & 0.1900 & 0.2199 & 0.2020 & 0.1187 & 0.0284 & 0.1856 & 0.0407 & 0.1930 & 0.0885 & 0.2297 \\
    TREC-Robust 
     & BLEU-1 & 0.0391 & 0.3135 & 0.1308 & 0.3400 & 0.2327 & 0.3605 & 0.3163 & 0.0905 & 0.0106 & 0.1803 & 0.0315 & 0.2921 & 0.0432 & 0.3408 \\
     & BLEU-2 & 0.0115 & 0.2286 & 0.0529 & 0.2528 & 0.1469 & 0.2628 & 0.2253 & 0.0278 & 0.0026 & 0.1090 & 0.0099 & 0.2197 & 0.0117 & 0.2564 \\
     & ROUGE-L & 0.0876 & 0.3377 & 0.0998 & 0.3703 & 0.2791 & 0.3606 & 0.3465 & 0.0905 & 0.0216 & 0.2537 & 0.0417 & 0.3508 & 0.0790 & 0.3597 \\
    TREC-COVID 
     & BLEU-1 & 0.0269 & 0.2747 & 0.0728 & 0.1879 & 0.1582 & 0.1188 & 0.1565 & 0.0859 & 0.0095 & 0.1359 & 0.0290 & 0.1648 & 0.0207 & 0.0828 \\
     & BLEU-2 & 0.0081 & 0.1599 & 0.0267 & 0.1022 & 0.0851 & 0.0391 & 0.0742 & 0.0182 & 0.0042 & 0.0372 & 0.0103 & 0.0619 & 0.0041 & 0.0233 \\
     & ROUGE-L & 0.0530 & 0.2310 & 0.0488 & 0.2371 & 0.2185 & 0.1633 & 0.2565 & 0.0904 & 0.0030 & 0.1429 & 0.0094 & 0.1735 & 0.0443 & 0.1111 \\
    FIRE
     & BLEU-1 & 0.0559 & 0.3745 & 0.1285 & 0.3400 & 0.3058 & 0.3122 & 0.3529 & 0.0943 & 0.0161 & 0.2694 & 0.0421 & 0.3125 & 0.0534 & 0.3420 \\
     & BLEU-2 & 0.0270 & 0.2615 & 0.0654 & 0.2308 & 0.2043 & 0.2058 & 0.2487 & 0.0583 & 0.0034 & 0.1749 & 0.0177 & 0.2321 & 0.0207 & 0.2253 \\
     & ROUGE-L & 0.0967 & 0.3522 & 0.1264 & 0.3422 & 0.3300 & 0.3224 & 0.3578 & 0.1335 & 0.0290 & 0.2796 & 0.0543 & 0.3492 & 0.1044 & 0.3495 \\
    \midrule
    \multicolumn{16}{l}{\textbf{{Query Expansion}}} \\
    GOV2 
     & BLEU-1 & 0.0303 & 0.3612 & 0.0751 & 0.3208 & 0.3398 & 0.3110 & 0.3006 & 0.0399 & 0.0035 & 0.2070 & 0.0187 & 0.2833 & 0.0278 & 0.2752 \\
     & BLEU-2 & 0.0170 & 0.2642 & 0.0369 & 0.2066 & 0.2297 & 0.2080 & 0.1957 & 0.0216 & 0.0017 & 0.0884 & 0.0097 & 0.1989 & 0.0112 & 0.1766 \\
     & ROUGE-L & 0.0672 & 0.4527 & 0.0954 & 0.4013 & 0.3891 & 0.3853 & 0.3736 & 0.1179 & 0.0085 & 0.2337 & 0.0165 & 0.3867 & 0.0539 & 0.3881 \\
    TREC-Robust
     & BLEU-1 & 0.0305 & 0.3807 & 0.0631 & 0.4260 & 0.3502 & 0.3333 & 0.4006 & 0.0768 & 0.0049 & 0.2483 & 0.0503 & 0.3020 & 0.0329 & 0.3621 \\
     & BLEU-2 & 0.0139 & 0.3100 & 0.0330 & 0.3410 & 0.2472 & 0.2335 & 0.3265 & 0.0415 & 0.0026 & 0.1682 & 0.0277 & 0.2237 & 0.0136 & 0.2993 \\
     & ROUGE-L & 0.0735 & 0.4681 & 0.1108 & 0.4543 & 0.4059 & 0.3621 & 0.4497 & 0.1348 & 0.0102 & 0.2636 & 0.0687 & 0.4206 & 0.0596 & 0.4466 \\
    TREC-COVID
     & BLEU-1 & 0.0141 & 0.1551 & 0.0475 & 0.2207 & 0.1531 & 0.1746 & 0.1531 & 0.0459 & 0.0069 & 0.0968 & 0.0169 & 0.1477 & 0.0257 & 0.1089 \\
     & BLEU-2 & 0.0045 & 0.0878 & 0.0174 & 0.1108 & 0.0774 & 0.1134 & 0.0774 & 0.0223 & 0.0031 & 0.0359 & 0.0085 & 0.0727 & 0.0126 & 0.0381 \\
     & ROUGE-L & 0.0551 & 0.3245 & 0.0424 & 0.3036 & 0.2912 & 0.3722 & 0.2984 & 0.1389 & 0.0490 & 0.1861 & 0.0217 & 0.2931 & 0.0454 & 0.2449 \\
    FIRE
     & BLEU-1 & 0.0326 & 0.4465 & 0.0510 & 0.3985 & 0.2739 & 0.3112 & 0.3257 & 0.0649 & 0.0102 & 0.1362 & 0.0302 & 0.4044 & 0.0237 & 0.3169 \\
     & BLEU-2 & 0.0183 & 0.3653 & 0.0313 & 0.3118 & 0.1845 & 0.2281 & 0.2442 & 0.0366 & 0.0056 & 0.0736 & 0.0152 & 0.3123 & 0.0110 & 0.2391 \\
     & ROUGE-L & 0.0686 & 0.5255 & 0.1270 & 0.4771 & 0.4097 & 0.4326 & 0.4487 & 0.1267 & 0.0094 & 0.2255 & 0.0618 & 0.4416 & 0.0468 & 0.4476 \\
    Query2Doc
     & BLEU-1 & 0.1065 & 0.3061 & 0.1952 & 0.3011 & 0.2984 & 0.3038 & 0.3032 & 0.1983 & 0.0169 & 0.1253 & 0.0739 & 0.3045 & 0.1262 & 0.2916 \\
     & BLEU-2 & 0.0512 & 0.1712 & 0.1055 & 0.1729 & 0.1762 & 0.1723 & 0.1732 & 0.1047 & 0.0066 & 0.0547 & 0.0312 & 0.1626 & 0.0626 & 0.1549 \\
     & ROUGE-L & 0.1417 & 0.2578 & 0.1554 & 0.2698 & 0.2525 & 0.2578 & 0.2622 & 0.1617 & 0.0199 & 0.1488 & 0.0643 & 0.2558 & 0.1537 & 0.2398 \\
    TREC-CAsT 
     & BLEU-1 & 0.0194 & 0.1806 & 0.0307 & 0.2295 & 0.2258 & 0.1477 & 0.2513 & 0.0937 & 0.0000 & 0.0343 & 0.0145 & 0.1718 & 0.0205 & 0.1974 \\
     & BLEU-2 & 0.0087 & 0.1205 & 0.0074 & 0.1568 & 0.1259 & 0.0705 & 0.1740 & 0.0554 & 0.0000 & 0.0180 & 0.0055 & 0.1006 & 0.0067 & 0.1291 \\
     & ROUGE-L & 0.0453 & 0.2138 & 0.0282 & 0.2691 & 0.2373 & 0.2037 & 0.2923 & 0.0614 & 0.0046 & 0.1858 & 0.0199 & 0.3265 & 0.0393 & 0.2963 \\
    TREC-Web
     & BLEU-1 & 0.0109 & 0.3596 & 0.0321 & 0.3652 & 0.3368 & 0.2100 & 0.3492 & 0.0313 & 0.0066 & 0.1318 & 0.0078 & 0.2627 & 0.0159 & 0.4944 \\
     & BLEU-2 & 0.0042 & 0.2998 & 0.0170 & 0.3016 & 0.2756 & 0.1649 & 0.2660 & 0.0180 & 0.0021 & 0.0661 & 0.0020 & 0.2181 & 0.0056 & 0.4587 \\
     & ROUGE-L & 0.0381 & 0.4967 & 0.0912 & 0.4738 & 0.4757 & 0.3889 & 0.4338 & 0.1761 & 0.0208 & 0.1967 & 0.0292 & 0.4692 & 0.0376 & 0.6337 \\
    \midrule
    \multicolumn{16}{l}{\textbf{Query Reformulation}} \\
    CODEC 
     & Precision & 0.0000 & 0.3333 & 0.0000 & 0.0729 & 0.0000 & 0.0357 & 0.0250 & 0.0000 & 0.0000 & 0.0000 & 0.0000 & 0.0357 & 0.0000 & 0.0278 \\
     & Recall & 0.0000 & 0.1333 & 0.0000 & 0.1333 & 0.0000 & 0.0833 & 0.0833 & 0.0000 & 0.0000 & 0.0000 & 0.0000 & 0.0833 & 0.0000 & 0.0833 \\
     & F1 & 0.0000 & 0.1667 & 0.0000 & 0.0940 & 0.0000 & 0.0500 & 0.0385 & 0.0000 & 0.0000 & 0.0000 & 0.0000 & 0.0500 & 0.0000 & 0.0417 \\
    QReCC 
     & BLEU-1 & 0.0389 & 0.7474 & 0.0795 & 0.7451 & 0.7347 & 0.7570 & 0.7611 & 0.2027 & 0.0102 & 0.6650 & 0.0410 & 0.7493 & 0.0248 & 0.7446 \\
     & BLEU-2 & 0.0303 & 0.6879 & 0.0575 & 0.6857 & 0.6675 & 0.6935 & 0.6973 & 0.1604 & 0.0066 & 0.5917 & 0.0310 & 0.6890 & 0.0173 & 0.6830 \\
     & ROUGE-L & 0.0836 & 0.8127 & 0.1431 & 0.8065 & 0.8011 & 0.8196 & 0.8174 & 0.4568 & 0.0257 & 0.7280 & 0.0916 & 0.8029 & 0.0541 & 0.8123 \\
    CANARD 
     & BLEU-1 & 0.0346 & 0.7523 & 0.0586 & 0.7497 & 0.7711 & 0.7434 & 0.7289 & 0.1814 & 0.0122 & 0.6920 & 0.0240 & 0.7524 & 0.0165 & 0.7235 \\
     & BLEU-2 & 0.0259 & 0.7020 & 0.0428 & 0.6994 & 0.7208 & 0.6907 & 0.6740 & 0.1355 & 0.0085 & 0.6348 & 0.0167 & 0.7006 & 0.0106 & 0.6695 \\
     & ROUGE-L & 0.0773 & 0.8371 & 0.0783 & 0.8342 & 0.8388 & 0.8259 & 0.8269 & 0.3085 & 0.0296 & 0.7867 & 0.0448 & 0.8328 & 0.0415 & 0.8150 \\
    TREC-CAsT 
     & BLEU-1 & 0.0365 & 0.7172 & 0.0781 & 0.6545 & 0.7188 & 0.6800 & 0.7300 & 0.1227 & 0.0084 & 0.6512 & 0.0222 & 0.6545 & 0.0123 & 0.6800 \\
     & BLEU-2 & 0.0254 & 0.6473 & 0.0589 & 0.6000 & 0.6464 & 0.6208 & 0.6740 & 0.0744 & 0.0032 & 0.5988 & 0.0136 & 0.5778 & 0.0067 & 0.6085 \\
     & ROUGE-L & 0.0846 & 0.7941 & 0.1156 & 0.7769 & 0.7982 & 0.7615 & 0.8030 & 0.2279 & 0.0159 & 0.7602 & 0.0297 & 0.7632 & 0.0320 & 0.7829 \\
    GECOR 
     & BLEU-1 & 0.0285 & 0.8918 & 0.0503 & 0.9118 & 0.8920 & 0.8789 & 0.9108 & 0.1550 & 0.0110 & 0.8142 & 0.0211 & 0.8670 & 0.0175 & 0.8879 \\
     & BLEU-2 & 0.0234 & 0.8594 & 0.0372 & 0.8789 & 0.8593 & 0.8446 & 0.8751 & 0.1232 & 0.0087 & 0.7644 & 0.0156 & 0.8361 & 0.0091 & 0.8512 \\
     & ROUGE-L & 0.0592 & 0.9544 & 0.0541 & 0.9579 & 0.9485 & 0.9365 & 0.9461 & 0.2568 & 0.0216 & 0.8860 & 0.0297 & 0.9400 & 0.0385 & 0.9429 \\
     \midrule
     \multicolumn{16}{l}{\textbf{Query Clarification}} \\
     MIMICS
     & Precision & 0.0000 & 0.2154 & 0.0000 & 0.2161 & 0.1442 & 0.1788 & 0.1985 & 0.0000 & 0.0000 & 0.1033 & 0.0000 & 0.1792 & 0.0000 & 0.1871 \\
     & Recall & 0.0000 & 0.2217 & 0.0000 & 0.2353 & 0.1609 & 0.1880 & 0.2139 & 0.0000 & 0.0000 & 0.1090 & 0.0000 & 0.1951 & 0.0000 & 0.2020 \\
     & F1 & 0.0000 & 0.2142 & 0.0000 & 0.2207 & 0.1492 & 0.1780 & 0.2007 & 0.0000 & 0.0000 & 0.1040 & 0.0000 & 0.1824 & 0.0000 & 0.1902 \\
    MIMICS-Duo
     & Precision & 0.0000 & 0.2665 & 0.0000 & 0.2676 & 0.2399 & 0.3125 & 0.2819 & 0.0000 & 0.0000 & 0.2725 & 0.0000 & 0.2473 & 0.0000 & 0.2870 \\
     & Recall & 0.0000 & 0.2700 & 0.0000 & 0.2934 & 0.2546 & 0.3288 & 0.2995 & 0.0000 & 0.0000 & 0.2934 & 0.0000 & 0.2733 & 0.0000 & 0.3090 \\
     & F1 & 0.0000 & 0.2570 & 0.0000 & 0.2654 & 0.2392 & 0.3113 & 0.2759 & 0.0000 & 0.0000 & 0.2698 & 0.0000 & 0.2446 & 0.0000 & 0.2848 \\
    ClariQ-FKw 
     & BLEU-1 & 0.0155 & 0.3603 & 0.0573 & 0.3531 & 0.3557 & 0.3390 & 0.3320 & 0.0598 & 0.0009 & 0.3514 & 0.0151 & 0.3546 & 0.0190 & 0.3664 \\
     & BLEU-2 & 0.0075 & 0.2510 & 0.0333 & 0.2420 & 0.2515 & 0.2299 & 0.2189 & 0.0267 & 0.0005 & 0.2446 & 0.0071 & 0.2426 & 0.0094 & 0.2602 \\
     & ROUGE-L & 0.0402 & 0.3631 & 0.0941 & 0.3551 & 0.3642 & 0.3464 & 0.3391 & 0.1786 & 0.0053 & 0.3602 & 0.0180 & 0.3599 & 0.0402 & 0.3681 \\
    RaoCQ 
     & BLEU-1 & 0.0131 & 0.1731 & 0.0223 & 0.1253 & 0.1755 & 0.1882 & 0.1451 & 0.0340 & 0.0039 & 0.1694 & 0.0070 & 0.1378 & 0.0133 & 0.1575 \\
     & BLEU-2 & 0.0037 & 0.0259 & 0.0062 & 0.0132 & 0.0265 & 0.0336 & 0.0261 & 0.0092 & 0.0015 & 0.0260 & 0.0026 & 0.0382 & 0.0045 & 0.0471 \\
     & ROUGE-L & 0.0235 & 0.1016 & 0.0150 & 0.1052 & 0.1085 & 0.0986 & 0.0962 & 0.0490 & 0.0054 & 0.1084 & 0.0016 & 0.1016 & 0.0273 & 0.0863 \\
    \midrule
    \multicolumn{16}{l}{\textbf{Query Subtopic Generation}} \\
    TREC-Web 
     & Precision & 0.0000 & 0.0754 & 0.0000 & 0.0888 & 0.0480 & 0.0623 & 0.0762 & 0.0000 & 0.0000 & 0.0480 & 0.0000 & 0.1000 & 0.0000 & 0.1080 \\
     & Recall & 0.0000 & 0.1360 & 0.0000 & 0.1620 & 0.0913 & 0.1327 & 0.1460 & 0.0000 & 0.0000 & 0.0533 & 0.0000 & 0.1460 & 0.0000 & 0.1560 \\
     & F1 & 0.0000 & 0.0892 & 0.0000 & 0.1076 & 0.0580 & 0.0778 & 0.0913 & 0.0000 & 0.0000 & 0.0497 & 0.0000 & 0.1089 & 0.0000 & 0.1189 \\
    \midrule
    \multicolumn{16}{l}{\textbf{Query Suggestion}} \\
    AOL 
     & BLEU-1 & 0.0082 & 0.3936 & 0.0192 & 0.4238 & 0.3878 & 0.3794 & 0.4140 & 0.0367 & 0.0031 & 0.3301 & 0.0089 & 0.3494 & 0.0032 & 0.3956 \\
     & BLEU-2 & 0.0054 & 0.2962 & 0.0111 & 0.3218 & 0.2877 & 0.2797 & 0.3157 & 0.0212 & 0.0021 & 0.2406 & 0.0056 & 0.2532 & 0.0018 & 0.3005 \\
     & ROUGE-L & 0.0198 & 0.4804 & 0.0383 & 0.5024 & 0.4673 & 0.5097 & 0.5087 & 0.2148 & 0.0045 & 0.4263 & 0.0681 & 0.5059 & 0.0085 & 0.4737 \\
    \midrule
    \multicolumn{16}{l}{\textbf{Query Matching}} \\
    MSRP 
     & Acc & 0.4250 & 0.8600 & 0.4400 & 0.8550 & 0.7850 & 0.8350 & 0.7950 & 0.6300 & 0.3250 & 0.7000 & 0.4000 & 0.8850 & 0.3900 & 0.7350 \\
     & F1 & 0.4387 & 0.8592 & 0.4587 & 0.8554 & 0.7856 & 0.8346 & 0.7955 & 0.6480 & 0.2675 & 0.6566 & 0.4049 & 0.8834 & 0.3950 & 0.7476 \\
    \midrule
    \multicolumn{16}{l}{\textbf{Query Intent Classification}} \\
    MANtIS 
     & P@1 & 0.0650 & 0.4550 & 0.1000 & 0.4750 & 0.4200 & 0.4250 & 0.4650 & 0.1650 & 0.0050 & 0.3750 & 0.0850 & 0.4400 & 0.1550 & 0.4300 \\
    ORCAS-I 
     & Acc & 0.3600 & 0.4900 & 0.3200 & 0.4700 & 0.4900 & 0.5000 & 0.4800 & 0.2200 & 0.1000 & 0.4900 & 0.2100 & 0.4600 & 0.3200 & 0.4500 \\
     & F1 & 0.2918 & 0.4206 & 0.2486 & 0.4084 & 0.4200 & 0.4329 & 0.4074 & 0.1905 & 0.0865 & 0.4265 & 0.1526 & 0.3959 & 0.2435 & 0.3820 \\
    TREC-Web
     & Acc & 0.2000 & 0.8500 & 0.3000 & 0.9000 & 0.3500 & 0.7500 & 0.7000 & 0.2500 & 0.4000 & 0.5500 & 0.2000 & 0.7500 & 0.8500 & 0.9000 \\
     & F1 & 0.2667 & 0.8726 & 0.3733 & 0.8863 & 0.4548 & 0.8296 & 0.7543 & 0.3200 & 0.4303 & 0.5881 & 0.2667 & 0.7712 & 0.8416 & 0.9000 \\
    \bottomrule
    \end{tabular}
    }
    \caption{Results for eight query understanding tasks. ``Vanilla'' denotes the model without fine-tuning. ``+25\%'' means using 25\% of \ourdata{} for training.}
    \label{tab:query_result}
\end{table*}

\begin{table*}[t]
    \centering
    \tiny
    \setlength{\tabcolsep}{1.3mm}{
    \begin{tabular}{llcccccccccccccc}
    \toprule
    &  & \multicolumn{2}{c}{LLaMA-2-Base} & \multicolumn{6}{c}{LLaMA-2-Chat} & \multicolumn{2}{c}{Falcon} & \multicolumn{2}{c}{Minima} & \multicolumn{2}{c}{Mistral} \\
    \cmidrule(lr){3-4}\cmidrule(lr){5-10}\cmidrule(lr){11-12}\cmidrule(lr){13-14}\cmidrule(lr){15-16}
    Task \& Dataset & Metric & Vanilla & +\ourdata{} & Vanilla & +\ourdata{} & +25\% & +50\% & +75\% & +FLAN & Vanilla & +\ourdata{} & Vanilla & +\ourdata{} & Vanilla & +\ourdata{} \\
    \midrule
    \multicolumn{16}{l}{\textbf{Fact Verification}} \\
    FEVER 
     & Acc & 0.6850 & 0.9050 & 0.6650 & 0.9300 & 0.9050 & 0.9450 & 0.9200 & 0.7150 & 0.6950 & 0.7550 & 0.6800 & 0.9250 & 0.7450 & 0.9000 \\
     & F1 & 0.6364 & 0.9000 & 0.6405 & 0.9295 & 0.9082 & 0.9444 & 0.9176 & 0.7159 & 0.6373 & 0.6812 & 0.6427 & 0.9237 & 0.6361 & 0.8993 \\
    Climate-FEVER 
     & Acc & 0.4248 & 0.5882 & 0.4771 & 0.5882 & 0.5882 & 0.6144 & 0.6078 & 0.3922 & 0.3595 & 0.3922 & 0.4248 & 0.5359 & 0.4379 & 0.5948 \\
     & F1 & 0.3177 & 0.5659 & 0.3163 & 0.5732 & 0.5792 & 0.5962 & 0.5865 & 0.3425 & 0.3240 & 0.3250 & 0.2910 & 0.5298 & 0.2997 & 0.5437 \\
    SciFact 
     & Acc & 0.4805 & 0.7922 & 0.4805 & 0.8182 & 0.6623 & 0.8182 & 0.8052 & 0.6883 & 0.7143 & 0.6883 & 0.4935 & 0.7792 & 0.6623 & 0.7662 \\
     & F1 & 0.5040 & 0.7555 & 0.5056 & 0.7860 & 0.6786 & 0.8123 & 0.7746 & 0.6452 & 0.5952 & 0.6023 & 0.5173 & 0.7741 & 0.6363 & 0.7249 \\
    \midrule
    \multicolumn{16}{l}{\textbf{Conversational QA}} \\
    CoQA & EM & 0.0032 & 0.3375 & 0.0064 & 0.3455 & 0.3100 & 0.3497 & 0.3141 & 0.0372 & 0.0000 & 0.0932 & 0.0010 & 0.3216 & 0.0061 & 0.3097 \\
    QuAC & EM & 0.0000 & 0.1735 & 0.0021 & 0.1924 & 0.1857 & 0.1954 & 0.1914 & 0.0082 & 0.0000 & 0.1308 & 0.0000 & 0.1998 & 0.0000 & 0.2194 \\
    \midrule
    \multicolumn{16}{l}{\textbf{Summarization}} \\
    CNN/DM 
     & ROUGE-1 & 0.2236 & 0.3640 & 0.2928 & 0.3852 & 0.3773 & 0.3666 & 0.3729 & 0.3285 & 0.0245 & 0.3083 & 0.0773 & 0.3679 & 0.2158 & 0.3743 \\
     & ROUGE-2 & 0.1016 & 0.1649 & 0.1117 & 0.1791 & 0.1646 & 0.1553 & 0.1654 & 0.1395 & 0.0011 & 0.1178 & 0.0339 & 0.1578 & 0.0921 & 0.1536 \\
     & ROUGE-L & 0.1481 & 0.2475 & 0.1852 & 0.2649 & 0.2560 & 0.2433 & 0.2497 & 0.2172 & 0.0224 & 0.2049 & 0.0570 & 0.2490 & 0.1431 & 0.2425 \\
    XSum
     & ROUGE-1 & 0.0968 & 0.3604 & 0.1428 & 0.3699 & 0.3562 & 0.3618 & 0.3577 & 0.2728 & 0.0177 & 0.2240 & 0.0421 & 0.3335 & 0.0846 & 0.3196 \\
     & ROUGE-2 & 0.0197 & 0.1401 & 0.0381 & 0.1469 & 0.1342 & 0.1373 & 0.1303 & 0.0997 & 0.0006 & 0.0459 & 0.0078 & 0.1173 & 0.0186 & 0.1040 \\
     & ROUGE-L & 0.0714 & 0.2840 & 0.1039 & 0.2866 & 0.2802 & 0.2811 & 0.2719 & 0.2087 & 0.0155 & 0.1672 & 0.0304 & 0.2592 & 0.0611 & 0.2399 \\
    WIkiSum 
     & ROUGE-1 & 0.1348 & 0.2709 & 0.1432 & 0.2776 & 0.2772 & 0.2788 & 0.2751 & 0.1358 & 0.0301 & 0.2326 & 0.0317 & 0.2566 & 0.1168 & 0.2935 \\
     & ROUGE-2 & 0.0363 & 0.1099 & 0.0410 & 0.1120 & 0.1151 & 0.1122 & 0.1115 & 0.0406 & 0.0014 & 0.0745 & 0.0064 & 0.1027 & 0.0250 & 0.1193 \\
     & ROUGE-L & 0.0849 & 0.1744 & 0.0885 & 0.1757 & 0.1765 & 0.1715 & 0.1729 & 0.0881 & 0.0293 & 0.1476 & 0.0217 & 0.1671 & 0.0771 & 0.1836 \\
    MultiNews 
     & ROUGE-1 & 0.1651 & 0.2226 & 0.1225 & 0.2201 & 0.2225 & 0.2252 & 0.2243 & 0.0950 & 0.0247 & 0.1627 & 0.0638 & 0.2050 & 0.1472 & 0.2256 \\
     & ROUGE-2 & 0.0565 & 0.0868 & 0.0361 & 0.0805 & 0.0885 & 0.0850 & 0.0887 & 0.0334 & 0.0018 & 0.0555 & 0.0185 & 0.0763 & 0.0455 & 0.0865 \\
     & ROUGE-L & 0.0918 & 0.1167 & 0.0670 & 0.1120 & 0.1188 & 0.1168 & 0.1205 & 0.0578 & 0.0228 & 0.0895 & 0.0371 & 0.1117 & 0.0793 & 0.1190 \\
    \midrule
    \multicolumn{16}{l}{\textbf{Reading Comprehension}} \\
    SQuAD 
     & F1 & 0.0448 & 0.7964 & 0.0124 & 0.8161 & 0.7873 & 0.7624 & 0.8279 & 0.7270 & 0.0427 & 0.5577 & 0.0390 & 0.7790 & 0.0599 & 0.7730 \\
    HotpotQA
     & F1 & 0.0396 & 0.8076 & 0.0943 & 0.8518 & 0.8219 & 0.7939 & 0.8435 & 0.4542 & 0.0181 & 0.4364 & 0.0386 & 0.8489 & 0.0449 & 0.8324 \\
    MS MARCO 
     & F1 & 0.1842 & 0.6601 & 0.3146 & 0.6575 & 0.6267 & 0.6279 & 0.6563 & 0.3693 & 0.1161 & 0.4877 & 0.0904 & 0.6473 & 0.1375 & 0.6567 \\
    BoolQ 
     & Acc & 0.6100 & 0.8150 & 0.6450 & 0.8400 & 0.7150 & 0.7550 & 0.8100 & 0.6500 & 0.4350 & 0.5750 & 0.5700 & 0.7700 & 0.5750 & 0.7050 \\
     & F1 & 0.5629 & 0.8162 & 0.5639 & 0.8425 & 0.7201 & 0.7580 & 0.8125 & 0.5613 & 0.4453 & 0.5580 & 0.4751 & 0.7740 & 0.5441 & 0.7104 \\
    WebGLM-QA 
     & BLEU1 & 0.2584 & 0.5153 & 0.1565 & 0.5223 & 0.5429 & 0.5249 & 0.5077 & 0.1186 & 0.0486 & 0.3896 & 0.0971 & 0.5470 & 0.2478 & 0.5192 \\
     & BLEU2 & 0.1782 & 0.4210 & 0.1048 & 0.4280 & 0.4325 & 0.4227 & 0.4149 & 0.0752 & 0.0252 & 0.2891 & 0.0629 & 0.4414 & 0.1683 & 0.4069 \\
     & ROUGE-L & 0.2362 & 0.4554 & 0.1566 & 0.4598 & 0.4414 & 0.4532 & 0.4588 & 0.1424 & 0.0539 & 0.3192 & 0.0819 & 0.4437 & 0.1764 & 0.4290 \\
    TriviaQA 
     & F1 & 0.0523 & 0.3932 & 0.0728 & 0.4019 & 0.4017 & 0.3857 & 0.3873 & 0.3232 & 0.0134 & 0.2742 & 0.0870 & 0.3410 & 0.0346 & 0.3436 \\
    \bottomrule
    \end{tabular}
    }
    \caption{Results for four document understanding tasks. ``Vanilla'' denotes the model without fine-tuning. ``+25\%'' means using 25\% of \ourdata{} for training.}
    \label{tab:doc_result}
\end{table*}

\begin{table*}[t]
    \centering
    \tiny
    \setlength{\tabcolsep}{1.3mm}{
    \begin{tabular}{llcccccccccccccc}
    \toprule
    &  & \multicolumn{2}{c}{LLaMA-2-Base} & \multicolumn{6}{c}{LLaMA-2-Chat} & \multicolumn{2}{c}{Falcon} & \multicolumn{2}{c}{Minima} & \multicolumn{2}{c}{Mistral} \\
    \cmidrule(lr){3-4}\cmidrule(lr){5-10}\cmidrule(lr){11-12}\cmidrule(lr){13-14}\cmidrule(lr){15-16}
    Task \& Dataset & Metric & Vanilla & +\ourdata{} & Vanilla & +\ourdata{} & +25\% & +50\% & +75\% & +FLAN & Vanilla & +\ourdata{} & Vanilla & +\ourdata{} & Vanilla & +\ourdata{} \\
    \midrule
    \multicolumn{16}{l}{\textbf{Passage Retrieval}} \\
    MS-MARCO 
     & MRR@10 & 0.0180 & 0.2407 & 0.0191 & 0.2416 & 0.2710 & 0.2537 & 0.2526 & 0.1953 & 0.0226 & 0.0147 & 0.0139 & 0.2394 & 0.0175 & 0.2464 \\
     & NDCG@10 & 0.0271 & 0.2971 & 0.0292 & 0.2985 & 0.3277 & 0.3106 & 0.3098 & 0.2457 & 0.0332 & 0.0218 & 0.0209 & 0.2957 & 0.0257 & 0.2996 \\
    \midrule
    \multicolumn{16}{l}{\textbf{Argument Retrieval}} \\
    Touché-2020 
     & MRR@10 & 0.1907 & 0.3147 & 0.1449 & 0.2037 & 0.3402 & 0.3455 & 0.3197 & 0.1904 & 0.2509 & 0.0850 & 0.1951 & 0.3210 & 0.0951 & 0.2255 \\
     & NDCG@10 & 0.0646 & 0.1565 & 0.0667 & 0.1132 & 0.1651 & 0.1560 & 0.1571 & 0.0732 & 0.1083 & 0.0410 & 0.0692 & 0.1549 & 0.0427 & 0.1283 \\
    ArguAna 
     & MRR@10 & 0.0265 & 0.1552 & 0.0082 & 0.2380 & 0.1823 & 0.2054 & 0.2397 & 0.0265 & 0.0161 & 0.0149 & 0.0401 & 0.1526 & 0.0026 & 0.1533 \\
     & NDCG@10 & 0.0441 & 0.2392 & 0.0144 & 0.3532 & 0.2809 & 0.3118 & 0.3585 & 0.0446 & 0.0255 & 0.0243 & 0.0641 & 0.2366 & 0.0042 & 0.2286 \\
    \midrule
    \multicolumn{16}{l}{\textbf{Bio-Medical IR}} \\
    TREC-Covid 
     & MRR@10 & 0.6012 & 0.8190 & 0.6209 & 0.8317 & 0.9233 & 0.8907 & 0.8862 & 0.7947 & 0.6081 & 0.6011 & 0.6415 & 0.8613 & 0.7021 & 0.8392 \\
     & NDCG@10 & 0.4011 & 0.5919 & 0.3849 & 0.6337 & 0.7203 & 0.6707 & 0.6526 & 0.5546 & 0.4141 & 0.3458 & 0.4015 & 0.6159 & 0.4096 & 0.6331 \\
    NFCorpus
     & MRR@10 & 0.2524 & 0.3984 & 0.3474 & 0.3766 & 0.5731 & 0.4887 & 0.4184 & 0.4572 & 0.2636 & 0.2610 & 0.2600 & 0.3294 & 0.2664 & 0.2624 \\
     & NDCG@10 & 0.1347 & 0.2321 & 0.1956 & 0.2272 & 0.3243 & 0.2779 & 0.2376 & 0.2694 & 0.1518 & 0.1419 & 0.1471 & 0.1925 & 0.1538 & 0.1595 \\
    \midrule
    \multicolumn{16}{l}{\textbf{Citation Prediction}} \\
    SciDocs 
     & MRR@10 & 0.0388 & 0.2950 & 0.0552 & 0.3004 & 0.3274 & 0.3102 & 0.3145 & 0.0878 & 0.0429 & 0.0407 & 0.0841 & 0.2715 & 0.0614 & 0.2079 \\
     & NDCG@10 & 0.0217 & 0.1628 & 0.0290 & 0.1671 & 0.1873 & 0.1749 & 0.1798 & 0.0479 & 0.0245 & 0.0206 & 0.0444 & 0.1537 & 0.0337 & 0.1117 \\
    \midrule
    \multicolumn{16}{l}{\textbf{Duplicate Question Retrieval}} \\
    Quora 
     & MRR@10 & 0.0295 & 0.8240 & 0.0331 & 0.8278 & 0.8406 & 0.8192 & 0.8070 & 0.2615 & 0.0373 & 0.0081 & 0.0579 & 0.7047 & 0.0238 & 0.8083 \\
     & NDCG@10 & 0.0368 & 0.8396 & 0.0406 & 0.8426 & 0.8533 & 0.8357 & 0.8258 & 0.2970 & 0.0474 & 0.0084 & 0.0754 & 0.3754 & 0.0290 & 0.8208 \\
    CQADupStack 
     & MRR@10 & 0.1566 & 0.3450 & 0.1309 & 0.3540 & 0.3612 & 0.3414 & 0.3366 & 0.1394 & 0.1437 & 0.1043 & 0.1486 & 0.3459 & 0.1498 & 0.3497 \\
     & NDCG@10 & 0.2019 & 0.3393 & 0.1846 & 0.3422 & 0.3497 & 0.3361 & 0.3331 & 0.1889 & 0.1941 & 0.1640 & 0.1963 & 0.3399 & 0.1986 & 0.3418 \\
    \midrule
    \multicolumn{16}{l}{\textbf{Entity Retrieval}} \\
    DBPedia 
     & MRR@10 & \multicolumn{1}{c}{0.2048} & 0.7262 & 0.1777 & 0.7386 & 0.7263 & 0.7314 & 0.7419 & 0.4663 & 0.1998 & 0.1449 & 0.1471 & 0.7047 & 0.1751 & 0.6906 \\
     & NDCG@10 & \multicolumn{1}{c}{0.0741} & 0.3961 & 0.0637 & 0.4026 & 0.4110 & 0.4030 & 0.4030 & 0.2237 & 0.0763 & 0.0568 & 0.0491 & 0.3754 & 0.0651 & 0.3697 \\
    \midrule
    \multicolumn{16}{l}{\textbf{Fact Checking}} \\
    FEVER
     & MRR@10 & 0.1214 & 0.8704 & 0.1303 & 0.8764 & 0.8336 & 0.8713 & 0.8800 & 0.2840 & 0.0781 & 0.0156 & 0.0156 & 0.8516 & 0.0306 & 0.8239 \\
     & NDCG@10 & 0.1487 & 0.8521 & 0.1775 & 0.8561 & 0.8232 & 0.8516 & 0.8587 & 0.3030 & 0.1093 & 0.0183 & 0.0248 & 0.8352 & 0.0419 & 0.8150 \\
    Climate-FEVER 
     & MRR@10 & 0.0709 & 0.3477 & 0.1093 & 0.3645 & 0.3005 & 0.3011 & 0.3351 & 0.0507 & 0.0258 & 0.0138 & 0.0180 & 0.2774 & 0.0102 & 0.2470 \\
     & NDCG@10 & 0.0589 & 0.2513 & 0.0876 & 0.2670 & 0.2193 & 0.2244 & 0.2490 & 0.0390 & 0.0214 & 0.0114 & 0.0154 & 0.1965 & 0.0086 & 0.1908 \\
    SciFact 
     & MRR@10 & 0.0132 & 0.7410 & 0.0133 & 0.7307 & 0.7143 & 0.6959 & 0.7204 & 0.1492 & 0.0193 & 0.0410 & 0.0221 & 0.6893 & 0.0222 & 0.6182 \\
     & NDCG@10 & 0.0217 & 0.7625 & 0.0224 & 0.7522 & 0.7343 & 0.7148 & 0.7491 & 0.2025 & 0.0274 & 0.0515 & 0.0401 & 0.7087 & 0.0339 & 0.6300 \\
    \midrule
    \multicolumn{16}{l}{\textbf{Question Answering}} \\
    NQ 
     & MRR@10 & 0.0207 & 0.4311 & 0.0316 & 0.4298 & 0.4471 & 0.4386 & 0.4548 & 0.2137 & 0.0253 & 0.0229 & 0.0208 & 0.4198 & 0.0157 & 0.3993 \\
     & NDCG@10 & 0.0302 & 0.4776 & 0.0438 & 0.4763 & 0.4906 & 0.4825 & 0.4952 & 0.2511 & 0.0360 & 0.0325 & 0.0298 & 0.4621 & 0.0219 & 0.4414 \\
    FiQA 
     & MRR@10 & 0.0244 & 0.4370 & 0.0437 & 0.4437 & 0.4757 & 0.4572 & 0.4282 & 0.1440 & 0.0367 & 0.0280 & 0.0288 & 0.3702 & 0.0365 & 0.3542 \\
     & NDCG@10 & 0.0204 & 0.3712 & 0.0355 & 0.3649 & 0.3911 & 0.3826 & 0.3591 & 0.1145 & 0.0321 & 0.0213 & 0.0267 & 0.3106 & 0.0369 & 0.2989 \\
    HotpotQA 
     & MRR@10 & 0.0380 & 0.8898 & 0.1018 & 0.8918 & 0.8515 & 0.8787 & 0.8916 & 0.2389 & 0.0489 & 0.0208 & 0.0427 & 0.8339 & 0.0342 & 0.8088 \\
     & NDCG@10 & 0.0388 & 0.7480 & 0.0955 & 0.7493 & 0.7154 & 0.7350 & 0.7510 & 0.2048 & 0.0470 & 0.0183 & 0.0423 & 0.6990 & 0.0338 & 0.6814 \\
    \bottomrule
    \end{tabular}
    }
    \caption{Results for eight query-document relationship understanding tasks. ``Vanilla'' denotes the model without fine-tuning. ``+25\%'' means using 25\% of \ourdata{} for training.}
    \label{tab:qd_result}
\end{table*}

\begin{table*}[t]
    \centering
    \tiny
    \begin{tabular}{llccccccccc}
    \toprule
    Task \& Dataset & Metric & \textit{w/o} Q & \textit{w/o} D & \textit{w/o} Q-D & \textit{w/o} QIC & \textit{w/o} FV & \textit{w/o} CP & \textit{w/o} Ds & \textit{w/o} Description & \textit{w/o} Template\\
    \midrule
    \multicolumn{11}{l}{\textbf{{Query Description}}} \\
    GOV2 
     & BLEU-1 & 0.0089 & 0.2343 & 0.2152 & 0.1990 & 0.1782 & 0.1324 & 0.1999 & 0.2077 & 0.1227 \\
     & BLEU-2 & 0.0037 & 0.1224 & 0.1062 & 0.0974 & 0.0849 & 0.0506 & 0.0992 & 0.0811 & 0.0570 \\
     & ROUGE-L & 0.0755 & 0.2051 & 0.2297 & 0.2037 & 0.2118 & 0.1166 & 0.2072 & 0.1787 & 0.1472 \\
    TREC-Robust 
     & BLEU-1 & 0.0624 & 0.2993 & 0.3552 & 0.3419 & 0.2919 & 0.1912 & 0.2129 & 0.2884 & 0.2303 \\
     & BLEU-2 & 0.0268 & 0.2084 & 0.2660 & 0.2611 & 0.2303 & 0.1098 & 0.0925 & 0.1955 & 0.1442 \\
     & ROUGE-L & 0.0852 & 0.3113 & 0.3616 & 0.3611 & 0.3647 & 0.1907 & 0.2218 & 0.2832 & 0.2043 \\
    TREC-COVID 
     & BLEU-1 & 0.0339 & 0.1418 & 0.1513 & 0.2412 & 0.1429 & 0.1474 & 0.1739 & 0.1057 & 0.0109 \\
     & BLEU-2 & 0.0176 & 0.0738 & 0.0786 & 0.1185 & 0.0758 & 0.0691 & 0.0894 & 0.0494 & 0.0048 \\
     & ROUGE-L & 0.1082 & 0.1830 & 0.2139 & 0.2896 & 0.1529 & 0.2399 & 0.1577 & 0.1977 & 0.0560 \\
    FIRE
     & BLEU-1 & 0.0044 & 0.3492 & 0.3087 & 0.3486 & 0.3015 & 0.1345 & 0.2967 & 0.1779 & 0.2085 \\
     & BLEU-2 & 0.0023 & 0.2342 & 0.2147 & 0.2311 & 0.2070 & 0.0893 & 0.2077 & 0.1179 & 0.1285 \\
     & ROUGE-L & 0.1191 & 0.3579 & 0.3547 & 0.3551 & 0.3684 & 0.1631 & 0.3691 & 0.1905 & 0.2388 \\
    \midrule
    \multicolumn{11}{l}{\textbf{{Query Expansion}}} \\
    GOV2 
     & BLEU-1 & 0.0828 & 0.2995 & 0.3311 & 0.2824 & 0.3284 & 0.3250 & 0.2892 & 0.3329 & 0.1656 \\
     & BLEU-2 & 0.0413 & 0.1768 & 0.2226 & 0.1977 & 0.2283 & 0.2296 & 0.2027 & 0.2342 & 0.0898 \\
     & ROUGE-L & 0.1321 & 0.3159 & 0.3830 & 0.3921 & 0.4384 & 0.3811 & 0.4132 & 0.4429 & 0.2544 \\
    TREC-Robust
     & BLEU-1 & 0.0460 & 0.3747 & 0.4179 & 0.4039 & 0.3915 & 0.3910 & 0.1809 & 0.3923 & 0.1180 \\
     & BLEU-2 & 0.0270 & 0.3067 & 0.3328 & 0.3383 & 0.3198 & 0.3264 & 0.1120 & 0.3181 & 0.0903 \\
     & ROUGE-L & 0.1027 & 0.4088 & 0.4421 & 0.4617 & 0.4409 & 0.4694 & 0.3274 & 0.4534 & 0.1744 \\
    TREC-COVID
     & BLEU-1 & 0.0719 & 0.1662 & 0.2839 & 0.1941 & 0.2295 & 0.2198 & 0.1729 & 0.1667 & 0.0891 \\
     & BLEU-2 & 0.0556 & 0.1113 & 0.1540 & 0.1163 & 0.1491 & 0.1238 & 0.0942 & 0.0957 & 0.0441 \\
     & ROUGE-L & 0.1879 & 0.3053 & 0.3424 & 0.2793 & 0.3659 & 0.3211 & 0.2950 & 0.2876 & 0.2275 \\
    FIRE
     & BLEU-1 & 0.1089 & 0.3871 & 0.2836 & 0.3993 & 0.4368 & 0.3311 & 0.4190 & 0.2947 & 0.2243 \\
     & BLEU-2 & 0.0514 & 0.2965 & 0.2227 & 0.3008 & 0.3449 & 0.2458 & 0.3285 & 0.2145 & 0.1647 \\
     & ROUGE-L & 0.1670 & 0.4313 & 0.4308 & 0.4867 & 0.4934 & 0.4147 & 0.4802 & 0.3784 & 0.2679 \\
    Query2Doc
     & BLEU-1 & 0.0378 & 0.3087 & 0.3278 & 0.2892 & 0.3215 & 0.2871 & 0.3262 & 0.2940 & 0.1278 \\
     & BLEU-2 & 0.0193 & 0.1712 & 0.1907 & 0.1639 & 0.1877 & 0.1632 & 0.1876 & 0.1680 & 0.0722 \\
     & ROUGE-L & 0.0819 & 0.2495 & 0.2781 & 0.2623 & 0.2642 & 0.2484 & 0.2724 & 0.2578 & 0.1251 \\
    TREC-CAsT 
     & BLEU-1 & 0.0016 & 0.1583 & 0.1817 & 0.1746 & 0.1546 & 0.1361 & 0.2000 & 0.1733 & 0.0008 \\
     & BLEU-2 & 0.0006 & 0.0972 & 0.1046 & 0.0950 & 0.1102 & 0.0880 & 0.1096 & 0.0999 & 0.0004 \\
     & ROUGE-L & 0.0480 & 0.2538 & 0.2058 & 0.2306 & 0.2184 & 0.2100 & 0.2922 & 0.1977 & 0.0320 \\
    TREC-Web
     & BLEU-1 & 0.0380 & 0.4848 & 0.2513 & 0.3871 & 0.4857 & 0.2996 & 0.4951 & 0.3333 & 0.1609 \\
     & BLEU-2 & 0.0125 & 0.4437 & 0.2089 & 0.3118 & 0.4203 & 0.2423 & 0.4337 & 0.2838 & 0.1104 \\
     & ROUGE-L & 0.1404 & 0.5897 & 0.4403 & 0.4774 & 0.5507 & 0.3782 & 0.5585 & 0.3889 & 0.2286 \\
    \midrule
    \multicolumn{11}{l}{\textbf{Query Reformulation}} \\
    CODEC 
     & Precision & 0.0000 & 0.0357 & 0.0590 & 0.1250 & 0.0694 & 0.0670 & 0.0313 & 0.0903 & 0.0000 \\
     & Recall & 0.0000 & 0.0833 & 0.1333 & 0.0833 & 0.1333 & 0.1333 & 0.0833 & 0.1333 & 0.0000 \\
     & F1 & 0.0000 & 0.0500 & 0.0812 & 0.1000 & 0.0913 & 0.0885 & 0.0455 & 0.0972 & 0.0000 \\
    QReCC 
     & BLEU-1 & 0.3044 & 0.7436 & 0.7490 & 0.7451 & 0.7557 & 0.7502 & 0.6801 & 0.7616 & 0.5111 \\
     & BLEU-2 & 0.2472 & 0.6858 & 0.6929 & 0.6832 & 0.6985 & 0.6901 & 0.6138 & 0.7030 & 0.4507 \\
     & ROUGE-L & 0.4010 & 0.8194 & 0.8180 & 0.8114 & 0.8250 & 0.8114 & 0.7502 & 0.8212 & 0.5914 \\
    CANARD 
     & BLEU-1 & 0.1128 & 0.7387 & 0.7456 & 0.7585 & 0.7505 & 0.7396 & 0.7456 & 0.7617 & 0.3375 \\
     & BLEU-2 & 0.0868 & 0.6860 & 0.6951 & 0.7073 & 0.6986 & 0.6899 & 0.6927 & 0.7091 & 0.2960 \\
     & ROUGE-L & 0.1704 & 0.8247 & 0.8272 & 0.8333 & 0.8348 & 0.8354 & 0.8333 & 0.8394 & 0.4082 \\
    TREC-CAsT 
     & BLEU-1 & 0.1253 & 0.6765 & 0.7389 & 0.6634 & 0.6887 & 0.7030 & 0.6979 & 0.6415 & 0.1440 \\
     & BLEU-2 & 0.1033 & 0.6063 & 0.6826 & 0.5791 & 0.6338 & 0.6399 & 0.6241 & 0.5722 & 0.1345 \\
     & ROUGE-L & 0.1933 & 0.7664 & 0.8062 & 0.7394 & 0.7851 & 0.7629 & 0.7683 & 0.7559 & 0.2545 \\
    GECOR 
     & BLEU-1 & 0.1470 & 0.8799 & 0.8834 & 0.9115 & 0.9004 & 0.8537 & 0.9056 & 0.8793 & 0.3524 \\
     & BLEU-2 & 0.1215 & 0.8450 & 0.8537 & 0.8795 & 0.8688 & 0.8206 & 0.8769 & 0.8485 & 0.3144 \\
     & ROUGE-L & 0.2149 & 0.9430 & 0.9536 & 0.9504 & 0.9485 & 0.9405 & 0.9563 & 0.9509 & 0.4089 \\
     \midrule
     \multicolumn{11}{l}{\textbf{Query Clarification}} \\
     MIMICS
     & Precision & 0.0000 & 0.2018 & 0.1940 & 0.2034 & 0.2038 & 0.1118 & 0.2072 & 0.1035 & 0.1967 \\
     & Recall & 0.0000 & 0.2149 & 0.2165 & 0.2107 & 0.2193 & 0.1245 & 0.2158 & 0.1138 & 0.2228 \\
     & F1 & 0.0000 & 0.2032 & 0.1989 & 0.2017 & 0.2065 & 0.1142 & 0.2072 & 0.1064 & 0.2042 \\
    MIMICS-Duo
     & Precision & 0.0000 & 0.2795 & 0.2998 & 0.2665 & 0.2861 & 0.1767 & 0.0471 & 0.1835 & 0.2718 \\
     & Recall & 0.0000 & 0.2886 & 0.3148 & 0.2786 & 0.3104 & 0.2000 & 0.0531 & 0.1903 & 0.2844 \\
     & F1 & 0.0000 & 0.2732 & 0.2924 & 0.2571 & 0.2842 & 0.1751 & 0.0460 & 0.1757 & 0.2675 \\
    ClariQ-FKw 
     & BLEU-1 & 0.0425 & 0.3547 & 0.3680 & 0.3500 & 0.3475 & 0.3183 & 0.3541 & 0.3197 & 0.0745 \\
     & BLEU-2 & 0.0263 & 0.2454 & 0.2587 & 0.2385 & 0.2397 & 0.2246 & 0.2461 & 0.2156 & 0.0418 \\
     & ROUGE-L & 0.1114 & 0.3585 & 0.3696 & 0.3575 & 0.3562 & 0.3383 & 0.3555 & 0.3329 & 0.1042 \\
    RaoCQ 
     & BLEU-1 & 0.0281 & 0.1989 & 0.1790 & 0.1506 & 0.1775 & 0.1630 & 0.1905 & 0.1811 & 0.0345 \\
     & BLEU-2 & 0.0083 & 0.0693 & 0.0449 & 0.0229 & 0.0413 & 0.0181 & 0.0480 & 0.0270 & 0.0054 \\
     & ROUGE-L & 0.0751 & 0.1275 & 0.1182 & 0.1035 & 0.1362 & 0.0918 & 0.1069 & 0.0927 & 0.0248 \\
    \midrule
    \multicolumn{11}{l}{\textbf{Query Subtopic Generation}} \\
    TREC-Web 
     & Precision & 0.0000 & 0.1134 & 0.0927 & 0.0728 & 0.0944 & 0.0463 & 0.0933 & 0.0333 & 0.0653 \\
     & Recall & 0.0000 & 0.1660 & 0.1220 & 0.1460 & 0.1427 & 0.1167 & 0.1240 & 0.0867 & 0.0960 \\
     & F1 & 0.0000 & 0.1241 & 0.1045 & 0.0916 & 0.1067 & 0.0641 & 0.1052 & 0.0468 & 0.0773 \\
    \midrule
    \multicolumn{11}{l}{\textbf{Query Suggestion}} \\
    AOL 
     & BLEU-1 & 0.0963 & 0.4273 & 0.4030 & 0.4443 & 0.3873 & 0.3981 & 0.4133 & 0.4466 & 0.3366 \\
     & BLEU-2 & 0.0623 & 0.3298 & 0.3022 & 0.3377 & 0.2872 & 0.2907 & 0.3147 & 0.3402 & 0.2479 \\
     & ROUGE-L & 0.2492 & 0.5152 & 0.4882 & 0.5313 & 0.4724 & 0.5008 & 0.4993 & 0.5175 & 0.4036 \\
    \midrule
    \multicolumn{11}{l}{\textbf{Query Matching}} \\
    MSRP 
     & Acc & 0.5500 & 0.8900 & 0.8400 & 0.8200 & 0.8800 & 0.7100 & 0.8350 & 0.7950 & 0.6000 \\
     & F1 & 0.5685 & 0.8900 & 0.8441 & 0.8200 & 0.8800 & 0.7241 & 0.8304 & 0.8018 & 0.6182 \\
    \midrule
    \multicolumn{11}{l}{\textbf{Query Intent Classification}} \\
    MANtIS 
     & P@1 & 0.1450 & 0.4450 & 0.4500 & 0.0750 & 0.4650 & 0.4450 & 0.4650 & 0.4700 & 0.0800 \\
    ORCAS-I 
     & Acc & 0.3700 & 0.4900 & 0.4900 & 0.3100 & 0.5000 & 0.5200 & 0.4600 & 0.5400 & 0.3800 \\
     & F1 & 0.3359 & 0.4131 & 0.4226 & 0.2806 & 0.4319 & 0.4577 & 0.3914 & 0.4704 & 0.3021 \\
    TREC-Web
     & Acc & 0.2000 & 0.8000 & 0.9000 & 0.4500 & 0.6000 & 0.8000 & 0.9000 & 0.9000 & 0.2000 \\
     & F1 & 0.2778 & 0.8433 & 0.9214 & 0.5143 & 0.6872 & 0.8133 & 0.9214 & 0.8863 & 0.2667 \\
    \bottomrule
    \end{tabular}
    \caption{Results for eight query understanding tasks on LLaMA-Chat with \ourdata{} removing different tasks/datasets. ``Q'' stands for ``query understanding'', ``D'' means ``document understanding'', ``Q-D'' means ``query-document relationship understanding'', ``QIC'' refers to ``query intent classification'', ``FV'' indicates ``fact verification'', and ``CP'' denotes ``citation prediction''.}
    \label{tab:query_result2}
\end{table*}

\begin{table*}[t]
    \centering
    \tiny
    \begin{tabular}{llccccccccc}
    \toprule
    Task \& Dataset & Metric & \textit{w/o} Q & \textit{w/o} D & \textit{w/o} Q-D & \textit{w/o} QIC & \textit{w/o} FV & \textit{w/o} CP & \textit{w/o} Ds & \textit{w/o} Description & \textit{w/o} Template\\
    \midrule
    \multicolumn{11}{l}{\textbf{Fact Verification}} \\
    FEVER 
     & Acc & 0.9250 & 0.6150 & 0.9200 & 0.9100 & 0.6800 & 0.9200 & 0.9500 & 0.8700 & 0.8600 \\
     & F1 & 0.9225 & 0.6185 & 0.9195 & 0.9066 & 0.6914 & 0.9195 & 0.9503 & 0.8729 & 0.8632 \\
    Climate-FEVER 
     & Acc & 0.6078 & 0.2288 & 0.5948 & 0.6013 & 0.2288 & 0.5752 & 0.4641 & 0.6013 & 0.4248 \\
     & F1 & 0.5838 & 0.2208 & 0.5743 & 0.5646 & 0.1351 & 0.5588 & 0.3903 & 0.5847 & 0.3628 \\
    SciFact 
     & Acc & 0.8052 & 0.6494 & 0.8312 & 0.8052 & 0.6753 & 0.8312 & 0.7013 & 0.6494 & 0.4935 \\
     & F1 & 0.7665 & 0.6267 & 0.8107 & 0.7816 & 0.6459 & 0.8047 & 0.7146 & 0.6646 & 0.5156 \\
    \midrule
    \multicolumn{11}{l}{\textbf{Conversational QA}} \\
    CoQA & EM & 0.3530 & 0.0000 & 0.3322 & 0.3285 & 0.3401 & 0.3350 & 0.3496 & 0.3504 & 0.0729 \\
    QuAC & EM & 0.1722 & 0.0000 & 0.1817 & 0.1632 & 0.2240 & 0.2051 & 0.2148 & 0.2141 & 0.0077 \\
    \midrule
    \multicolumn{11}{l}{\textbf{Summarization}} \\
    CNN/DM 
     & ROUGE-1 & 0.3737 & 0.2568 & 0.3777 & 0.3557 & 0.3869 & 0.3845 & 0.3840 & 0.3923 & 0.3450 \\
     & ROUGE-2 & 0.1679 & 0.1024 & 0.1726 & 0.1615 & 0.1676 & 0.1707 & 0.1675 & 0.1795 & 0.1568 \\
     & ROUGE-L & 0.2505 & 0.1768 & 0.2590 & 0.2413 & 0.2600 & 0.2576 & 0.2588 & 0.2685 & 0.2307 \\
    XSum
     & ROUGE-1 & 0.3645 & 0.1584 & 0.3765 & 0.3607 & 0.3669 & 0.3574 & 0.1756 & 0.3723 & 0.1790 \\
     & ROUGE-2 & 0.1372 & 0.0414 & 0.1554 & 0.1394 & 0.1501 & 0.1367 & 0.0460 & 0.1343 & 0.0679 \\
     & ROUGE-L & 0.2873 & 0.1294 & 0.3050 & 0.2799 & 0.2907 & 0.2824 & 0.1231 & 0.2868 & 0.1364 \\
    WIkiSum 
     & ROUGE-1 & 0.2783 & 0.1062 & 0.2788 & 0.2737 & 0.2807 & 0.3034 & 0.2773 & 0.3040 & 0.2734 \\
     & ROUGE-2 & 0.1145 & 0.0417 & 0.1149 & 0.1095 & 0.1155 & 0.1261 & 0.1148 & 0.1230 & 0.1109 \\
     & ROUGE-L & 0.1741 & 0.0749 & 0.1781 & 0.1759 & 0.1796 & 0.1946 & 0.1764 & 0.1912 & 0.1749 \\
    MultiNews 
     & ROUGE-1 & 0.2191 & 0.0877 & 0.2275 & 0.2191 & 0.2241 & 0.2363 & 0.2237 & 0.2352 & 0.1749 \\
     & ROUGE-2 & 0.0826 & 0.0275 & 0.0922 & 0.0801 & 0.0868 & 0.0891 & 0.0912 & 0.0942 & 0.0690 \\
     & ROUGE-L & 0.1142 & 0.0524 & 0.1239 & 0.1135 & 0.1216 & 0.1203 & 0.1217 & 0.1261 & 0.0936 \\
    \midrule
    \multicolumn{11}{l}{\textbf{Reading Comprehension}} \\
    SQuAD 
     & F1 & 0.8225 & 0.1075 & 0.8324 & 0.7940 & 0.7908 & 0.7849 & 0.8492 & 0.7735 & 0.1880 \\
    HotpotQA
     & F1 & 0.8518 & 0.1570 & 0.8753 & 0.8570 & 0.8643 & 0.8303 & 0.8823 & 0.8439 & 0.3230 \\
    MS MARCO 
     & F1 & 0.6872 & 0.2781 & 0.6430 & 0.6716 & 0.6727 & 0.6759 & 0.6502 & 0.6720 & 0.2286 \\
    BoolQ 
     & Acc & 0.8250 & 0.6000 & 0.7950 & 0.8300 & 0.8550 & 0.8450 & 0.8300 & 0.7900 & 0.5900 \\
     & F1 & 0.8274 & 0.5387 & 0.7947 & 0.8277 & 0.8563 & 0.8460 & 0.8330 & 0.7933 & 0.5970 \\
    WebGLM-QA 
     & BLEU1 & 0.5067 & 0.0002 & 0.5141 & 0.4977 & 0.5413 & 0.5090 & 0.5546 & 0.5081 & 0.0087 \\
     & BLEU2 & 0.4132 & 0.0002 & 0.4173 & 0.4048 & 0.4399 & 0.4139 & 0.4471 & 0.4082 & 0.0065 \\
     & ROUGE-L & 0.4520 & 0.0647 & 0.4481 & 0.4468 & 0.4623 & 0.4547 & 0.4580 & 0.4487 & 0.0919 \\
    TriviaQA 
     & F1 & 0.3919 & 0.1173 & 0.3892 & 0.3809 & 0.4068 & 0.3858 & 0.4038 & 0.3809 & 0.2187 \\
    \bottomrule
    \end{tabular}
    \caption{Results for four document understanding tasks on LLaMA-Chat with \ourdata{} removing different tasks/datasets. ``Q'' stands for ``query understanding'', ``D'' means ``document understanding'', ``Q-D'' means ``query-document relationship understanding'', ``QIC'' refers to ``query intent classification'', ``FV'' indicates ``fact verification'', and ``CP'' denotes ``citation prediction''.}
    \label{tab:doc_result2}
\end{table*}

\begin{table*}[t]
    \centering
    \tiny
    \begin{tabular}{llccccccccc}
    \toprule
    Task \& Dataset & Metric & \textit{w/o} Q & \textit{w/o} D & \textit{w/o} Q-D & \textit{w/o} QIC & \textit{w/o} FV & \textit{w/o} CP & \textit{w/o} Ds & \textit{w/o} Description & \textit{w/o} Template\\
    \midrule
    \multicolumn{11}{l}{\textbf{Passage Retrieval}} \\
    MS-MARCO 
     & MRR@10 & 0.2544 & 0.2365 & 0.1805 & 0.2482 & 0.2557 & 0.2201 & 0.2360 & 0.2424 & 0.1376 \\
     & NDCG@10 & 0.3090 & 0.2913 & 0.2254 & 0.3058 & 0.3114 & 0.2772 & 0.2934 & 0.2986 & 0.1813 \\
    \midrule
    \multicolumn{11}{l}{\textbf{Argument Retrieval}} \\
    Touché-2020 
     & MRR@10 & 0.3311 & 0.2313 & 0.4368 & 0.3251 & 0.2658 & 0.2093 & 0.2876 & 0.2068 & 0.2775 \\
     & NDCG@10 & 0.1641 & 0.1197 & 0.1754 & 0.1603 & 0.1361 & 0.1063 & 0.1393 & 0.0936 & 0.1096 \\
    ArguAna 
     & MRR@10 & 0.2768 & 0.2866 & 0.0365 & 0.2428 & 0.2298 & 0.1994 & 0.2449 & 0.1690 & 0.1629 \\
     & NDCG@10 & 0.4014 & 0.4044 & 0.0574 & 0.3621 & 0.3407 & 0.3078 & 0.3594 & 0.2640 & 0.2548 \\
    \midrule
    \multicolumn{11}{l}{\textbf{Bio-Medical IR}} \\
    TREC-Covid 
     & MRR@10 & 0.7920 & 0.8065 & 0.8385 & 0.8907 & 0.8638 & 0.8240 & 0.8198 & 0.8829 & 0.6506 \\
     & NDCG@10 & 0.5812 & 0.5928 & 0.5757 & 0.6593 & 0.6147 & 0.6292 & 0.6412 & 0.6285 & 0.4619 \\
    NFCorpus
     & MRR@10 & 0.4425 & 0.4818 & 0.5347 & 0.5074 & 0.3605 & 0.3692 & 0.3263 & 0.3290 & 0.3319 \\
     & NDCG@10 & 0.2758 & 0.2903 & 0.3148 & 0.3107 & 0.2059 & 0.2249 & 0.2026 & 0.2002 & 0.1941 \\
    \midrule
    \multicolumn{11}{l}{\textbf{Citation Prediction}} \\
    SciDocs 
     & MRR@10 & 0.3211 & 0.2646 & 0.2452 & 0.3147 & 0.3276 & 0.2878 & 0.3475 & 0.2989 & 0.1965 \\
     & NDCG@10 & 0.1827 & 0.1517 & 0.1366 & 0.1802 & 0.1866 & 0.1603 & 0.1993 & 0.1633 & 0.1097 \\
    \midrule
    \multicolumn{11}{l}{\textbf{Duplicate Question Retrieval}} \\
    Quora 
     & MRR@10 & 0.8259 & 0.7926 & 0.6551 & 0.7793 & 0.8062 & 0.7861 & 0.7438 & 0.8231 & 0.6882 \\
     & NDCG@10 & 0.8411 & 0.8135 & 0.6910 & 0.8060 & 0.8272 & 0.8079 & 0.7768 & 0.8383 & 0.7146 \\
    CQADupStack 
     & MRR@10 & 0.3458 & 0.3263 & 0.2487 & 0.3328 & 0.3358 & 0.3340 & 0.3496 & 0.3316 & 0.2429 \\
     & NDCG@10 & 0.3365 & 0.3232 & 0.2666 & 0.3306 & 0.3326 & 0.3283 & 0.3392 & 0.3266 & 0.2665 \\
    \midrule
    \multicolumn{11}{l}{\textbf{Entity Retrieval}} \\
    DBPedia 
     & MRR@10 & 0.7157 & 0.7053 & 0.5585 & 0.7329 & 0.7300 & 0.6734 & 0.7267 & 0.7239 & 0.5552 \\
     & NDCG@10 & 0.3955 & 0.3793 & 0.2974 & 0.4041 & 0.4115 & 0.3695 & 0.4084 & 0.4090 & 0.2823 \\
    \midrule
    \multicolumn{11}{l}{\textbf{Fact Checking}} \\
    FEVER
     & MRR@10 & 0.8732 & 0.8671 & 0.3111 & 0.8845 & 0.8794 & 0.8658 & 0.8698 & 0.8753 & 0.7681 \\
     & NDCG@10 & 0.8547 & 0.8485 & 0.3288 & 0.8618 & 0.8601 & 0.8495 & 0.8524 & 0.8554 & 0.7644 \\
    Climate-FEVER 
     & MRR@10 & 0.3130 & 0.3469 & 0.1360 & 0.3628 & 0.3869 & 0.3140 & 0.3855 & 0.2771 & 0.1100 \\
     & NDCG@10 & 0.2362 & 0.2476 & 0.1013 & 0.2647 & 0.2797 & 0.2340 & 0.2780 & 0.2049 & 0.0842 \\
    SciFact 
     & MRR@10 & 0.7265 & 0.6951 & 0.3260 & 0.7143 & 0.7648 & 0.6611 & 0.7378 & 0.7039 & 0.1721 \\
     & NDCG@10 & 0.7540 & 0.7276 & 0.3488 & 0.7512 & 0.7872 & 0.6996 & 0.7668 & 0.7289 & 0.2046 \\
    \midrule
    \multicolumn{11}{l}{\textbf{Question Answering}} \\
    NQ 
     & MRR@10 & 0.4472 & 0.4239 & 0.2951 & 0.4452 & 0.4803 & 0.4164 & 0.4295 & 0.4638 & 0.3261 \\
     & NDCG@10 & 0.4868 & 0.4647 & 0.3398 & 0.4867 & 0.5150 & 0.4617 & 0.4697 & 0.5003 & 0.3736 \\
     & MRR@10 & 0.4663 & 0.4325 & 0.2569 & 0.4157 & 0.4724 & 0.3845 & 0.4693 & 0.4249 & 0.3094 \\
     & NDCG@10 & 0.3838 & 0.3553 & 0.2068 & 0.3480 & 0.3923 & 0.3277 & 0.3809 & 0.3541 & 0.2708 \\
    HotpotQA 
     & MRR@10 & 0.8787 & 0.8558 & 0.3223 & 0.8787 & 0.8954 & 0.8548 & 0.8827 & 0.8975 & 0.7929 \\
     & NDCG@10 & 0.7427 & 0.7195 & 0.2702 & 0.7383 & 0.7565 & 0.7275 & 0.7477 & 0.7562 & 0.6777 \\
    \bottomrule
    \end{tabular}
    \caption{Results for eight query-document relationship understanding tasks on LLaMA-Chat with \ourdata{} removing different tasks/datasets. ``Q'' stands for ``query understanding'', ``D'' means ``document understanding'', ``Q-D'' means ``query-document relationship understanding'', ``QIC'' refers to ``query intent classification'', ``FV'' indicates ``fact verification'', and ``CP'' denotes ``citation prediction''.}
    \label{tab:qd_result2}
\end{table*}

\section{Data Examples}\label{app:example}
We present an example per task in Table~\ref{tab:exp1} -- Table~\ref{tab:exp12}.

\begin{table*}[t]
    \centering
    \small
    \begin{tabular}{p{.95\linewidth}}
    \toprule
        \textbf{Query Description}  \\
    \midrule
        \textbf{\texttt{Input}} \\
        The query description task involves describing documents potentially relevant to a user-provided query. Queries typically comprise keywords reflecting the user's information needs. The objective of the task is to articulate the characteristics and content of documents that would be considered pertinent to these keywords, aiding in the understanding and retrieval of relevant information.
        \\
        \\
        Query: Bhopal gas tragedy
        \\
        \\
        What are the relevant documents?  \\
        \midrule
        \textbf{\texttt{Target}} \\
        Relevant documents should contain information about the controversies ensuing after the verdict was passed on the long drawn legal tussle related to the Bhopal gas tragedy. The people's response to the verdict is valuable information too. \\
    \bottomrule
    \end{tabular}
    \caption{A data example of the query description task. Data source: FIRE.}
    \label{tab:exp1}
\end{table*}

\begin{table*}[t]
    \centering
    \small
    \begin{tabular}{p{.95\linewidth}}
    \toprule
        \textbf{Query Expansion}  \\
    \midrule
        \textbf{\texttt{Input}} \\
        The query expansion task involves elaborating an original, brief query into a longer, more detailed version while preserving the original search intent. This process enhances the search engine's understanding of the user's needs, leading to more accurate and relevant document retrieval.
        \\
        \\
        What is the expansion of this query:
        \\
        \\what does cybercrime mean
        \\
        \midrule
        \textbf{\texttt{Target}} \\
        Cybercrime is a broad term that refers to any criminal activity that involves the use of computer technology. It can include anything from fraudulent bank transfers to identity theft, posting of threatening messages online and distribution of malicious software. Cybercriminals use sophisticated software and hardware to commit their crimes, and their targets can range from individuals to large companies and organizations.\\
    \bottomrule
    \end{tabular}
    \caption{A data example of the query expansion task. Data source: Query2Doc.}
    \label{tab:exp2}
\end{table*}

\begin{table*}[t]
    \centering
    \small
    \begin{tabular}{p{.95\linewidth}}
    \toprule
        \textbf{Query Reformulation}  \\
    \midrule
        \textbf{\texttt{Input}} \\
        The query reformulation task enhances user-input queries to be more explicit and comprehensible for search engines. It addresses omissions typical of user queries, which often exclude common sense or contextually implied information. The refined query, therefore, includes all necessary details to guide the search engine towards retrieving the most relevant documents. \\
        \\
        The question is ``How much of a threat are ETFs to actively-managed Asset Managers?''. How can we rewrite the question in a new way? \\
        \midrule
        \textbf{\texttt{Target}} \\
        Reformulations: [1] Why do actively managed funds rarely beat the market? [2] How much of a threat are ETFs to actively-managed Asset Managers? [3] Passive vs active managed funds\\
    \bottomrule
    \end{tabular}
    \caption{A data example of the query reformulation task. Data source: CODEC.}
    \label{tab:exp3}
\end{table*}

\begin{table*}[t]
    \centering
    \small
    \begin{tabular}{p{.95\linewidth}}
    \toprule
        \textbf{Query Clarification}\\
    \midrule
        \textbf{\texttt{Input}} \\
        The query clarification task addresses unclear or ambiguous user queries by asking for further details or providing clarification options. This process helps refine the query, resulting in clearer and more precise search terms for improved search engine results.
        \\
        \\
        Considering the query ``what did the emancipation proclamation do'', what clarification options would you suggest? \\
    \midrule
        \textbf{\texttt{Target}} \\
        {[1]} how did {[2]} who did \\
    \bottomrule
    \end{tabular}
    \caption{A data example of the query clarification task. Data source: MIMICS.}
    \label{tab:exp4}
\end{table*}

\begin{table*}[t]
    \centering
    \small
    \begin{tabular}{p{.95\linewidth}}
    \toprule
        \textbf{Query Subtopic Generation} \\
    \midrule
        \textbf{\texttt{Input}} \\
        The query subtopic generation task addresses the ambiguity of web searches by identifying and presenting various aspects of the initial query. This approach aids search engines in understanding the query's breadth, leading to more diverse and relevant search results.\\\\Here are some descriptions about the query ``wind power'': What are the pros and cons of using wind power. Generate the query's subtopics.\\
        \midrule
        \textbf{\texttt{Target}} \\
        Subtopics: (1) What are the pros and cons of using wind power. (2) Find information on wind power in the USA. (3) Find information on wind power companies. (4) Find information on residential (home) wind power. (5) Find information on how wind turbines work. (6) Find pictures of wind turbines used for wind power. (7) Find pictures of a wind farm.\\
    \bottomrule
    \end{tabular}
    \caption{A data example of the query subtopic generation task. Data source: TREC-Web.}
    \label{tab:exp5}
\end{table*}

\begin{table*}[t]
    \centering
    \small
    \begin{tabular}{p{.95\linewidth}}
    \toprule
        \textbf{Query Suggestion} \\
    \midrule
        \textbf{\texttt{Input}} \\
        In search sessions, users often input a series of queries to fulfill a specific information need. The query suggestion task aims to analyze these queries and associated search behaviors to understand the user's intent and predict the next likely query, thereby enhancing the search experience.\\
        \\The search context is presented below: \\
        Query: tickets for nba draft \\
        Document title: tickets com online ticket broker selling tickets for concerts sports and theater events. \\
        Can you predict the next query?\\
        \midrule
        \textbf{\texttt{Target}} \\
        nba draft tickets\\
    \bottomrule
    \end{tabular}
    \caption{A data example of the query suggestion task. Data source: AOL.}
    \label{tab:exp6}
\end{table*}

\begin{table*}[t]
    \centering
    \small
    \begin{tabular}{p{.95\linewidth}}
    \toprule
        \textbf{Query Intent Classification} \\
    \midrule
        \textbf{\texttt{Input}} \\
        User queries can have various search intents, such as informational (seeking knowledge about a topic), transactional (aiming to purchase a product), or navigational (looking to find a specific website). Accurately discerning the type of intent behind a query is crucial for search engines to tailor and refine their results effectively.\\
        \\
        ``voting locations by zip code''\\
        What is the intent type of the query? Select one from the following options:\\
        (A) factual\\
        (B) abstain\\
        (C) instrumental\\
        (D) transactional\\
        (E) navigational\\
        \midrule
        \textbf{\texttt{Target}} \\
        factual\\
    \bottomrule
    \end{tabular}
    \caption{A data example of the query intent classification task. Data source: ORCAS-I.}
    \label{tab:exp7}
\end{table*}

\begin{table*}[t]
    \centering
    \small
    \begin{tabular}{p{.95\linewidth}}
    \toprule
        \textbf{Query Matching} \\
    \midrule
        \textbf{\texttt{Input}} \\
        The query matching task involves determining whether two queries or texts, despite differing in expression, convey the same meaning. This is crucial in search tasks where identifying synonymous queries can enhance the relevance and accuracy of results. 
        \\
        \\
        The driver, Eugene Rogers, helped to remove children from the bus, Wood said.
        \\
        \\
        At the accident scene, the driver was ``covered in blood'' but helped to remove children, Wood said.
        \\
        \\Do the above sentences mean the same thing? \\
        \midrule
        \textbf{\texttt{Target}} \\
        no\\
    \bottomrule
    \end{tabular}
    \caption{A data example of the query matching task. Data source: MSRP.}
    \label{tab:exp8}
\end{table*}

\begin{table*}[t]
    \centering
    \small
    \begin{tabular}{p{.95\linewidth}}
    \toprule
        \textbf{Fact Verification} \\
    \midrule
        \textbf{\texttt{Input}} \\
        The fact verification task is to assess whether a claim is supported or refuted by the given evidence. It requires a clear analysis of the relationship between the claim and the evidence, with careful examination to determine if there is enough information for making a judgment. It aids search engines in achieving a deeper comprehension of the documents.\\\\
        ``U2 is a Scottish rock band.''\\
        Based on ``(1) Title: U2 Content: U2 are an Irish rock band from Dublin formed in 1976 . (2) Title: U2 Content: Within four years , they signed with Island Records and released their debut album , Boy 1980 . (3) Title: U2 Content: Subsequent work such as their first UK number-one album , War 1983 , and the singles `` Sunday Bloody Sunday '' and `` Pride In the Name of Love '' helped establish U2 's reputation as a politically and socially conscious group . (4) Title: U2 Content: The group 's fifth album , The Joshua Tree 1987 , made them international superstars and was their greatest critical and commercial success . (5) Title: U2 Content: Topping music charts around the world , it produced their only number-one singles in the US , `` With or Without You '' and `` I Still Haven't Found What I 'm Looking For '' . (6) Title: U2 Content: Beginning with their acclaimed seventh album , Achtung Baby 1991 , and the multimedia intensive Zoo TV Tour , the band integrated influences from alternative rock , electronic dance music , and industrial music into their sound , and embraced a more ironic , flippant image . (7) Title: U2 Content: This experimentation continued through their ninth album , Pop 1997 , and the PopMart Tour , which were mixed successes . (8) Title: U2 Content: U2 regained critical and commercial favour with the records All That You Ca n't Leave Behind 2000 and How to Dismantle an Atomic Bomb 2004 , which established a more conventional , mainstream sound for the group . (9) Title: U2 Content: The group 's thirteenth album , Songs of Innocence 2014 , was released at no cost through the iTunes Store , but received criticism for its automatic placement in users' music libraries.'', which label {support or refute} should be assigned?\\
        \midrule
        \textbf{\texttt{Target}} \\
        refute\\
    \bottomrule
    \end{tabular}
    \caption{A data example of the query matching task. Data source: FEVER.}
    \label{tab:exp9}
\end{table*}

\begin{table*}[t]
    \centering
    \small
    \begin{tabular}{p{.95\linewidth}}
    \toprule
        \textbf{Conversational QA} \\
    \midrule
        \textbf{\texttt{Input}} \\
        Conversational question-answering involves responding to a series of interrelated questions based on a given context. As these questions might build upon shared information, some details may be implicitly understood rather than explicitly stated. By comprehensively understanding and analyzing this dialogue structure, search engines can enhance their interpretation of user queries and their connections to relevant documents, thereby improving result accuracy and relevance.\\
        \\
        In the context provided, answer the following questions:\\
        Malawi (, or ; or [maláwi]), officially the Republic of Malawi, is a landlocked country in southeast Africa that was formerly known as Nyasaland. It is bordered by Zambia to the northwest, Tanzania to the northeast, and Mozambique on the east, south and west. Malawi is over with an estimated population of 16,777,547 (July 2013 est.). Its capital is Lilongwe, which is also Malawi's largest city; the second largest is Blantyre, the third is Mzuzu and the fourth largest is its old capital Zomba. The name Malawi comes from the Maravi, an old name of the Nyanja people that inhabit the area. The country is also nicknamed ``The Warm Heart of Africa''. \\
        \\
        Malawi is among the smallest countries in Africa. Lake Malawi takes up about a third of Malawi's area. \\\\
        The area of Africa now known as Malawi was settled by migrating Bantu groups around the 10th century. Centuries later in 1891 the area was colonised by the British. In 1953 Malawi, then known as Nyasaland, a protectorate of the United Kingdom, became a protectorate within the semi-independent Federation of Rhodesia and Nyasaland. The Federation was dissolved in 1963. In 1964 the protectorate over Nyasaland was ended and Nyasaland became an independent country under Queen Elizabeth II with the new name Malawi. Two years later it became a republic. Upon gaining independence it became a one-party state under the presidency of Hastings Banda, who remained president until 1994, when he lost an election. Arthur Peter Mutharika is the current president. Malawi has a democratic, multi-party government. The country has a Malawian Defence Force that includes an army, a navy and an air wing. Malawi's foreign policy is pro-Western and includes positive diplomatic relations with most countries and participation in several international organisations, including the United Nations, the Commonwealth of Nations, the Southern African Development Community (SADC), the Common Market for Eastern and Southern Africa (COMESA), and the African Union (AU).\\
        Questions: (1) Who settled in Malawi? (2) when? (3) What does the defence force include? (4) What does COMESA stand for? (5) Is it a large country? (6) When was it colonized? (7) By who? (8) What was it called before Malawi? (9) when was the federation dissolved? (10) What is the capital city? (11) What is the biggest lake? (12) how much of Malawi's space does it take up? (13) What is AU?\\
        \midrule
        \textbf{\texttt{Target}} \\
        (1) Bantu groups (2) 10th century (3) army, a navy and an air wing (4) Common Market for Eastern and Southern Africa (5) No (6) 1891 (7) British (8) Nyasaland (9) 1963 (10) Lilongwe (11) Lake Malawi (12) a third of Malawi's area (13) African Union\\
    \bottomrule
    \end{tabular}
    \caption{A data example of the conversational QA task. Data source: CoQA.}
    \label{tab:exp10}
\end{table*}

\begin{table*}[t]
    \centering
    \small
    \begin{tabular}{p{.95\linewidth}}
    \toprule
        \textbf{Reading Comprehension} \\
    \midrule
        \textbf{\texttt{Input}} \\
        The reading comprehension task requires generating an answer to a question using the information from a given context. It necessitates a deep understanding of the text's context and semantics, enabling search engines to more accurately rank the relevance of retrieved documents based on this nuanced comprehension.\\
        \\
        Question: Who was the county where Jed Prouty Tavern and Inn is located named after?\\
        Consult the provided facts and generate your answer\\
        \\
        1. The Jed Prouty Tavern and Inn is an historic building at 57 Main Street in downtown Bucksport, Hancock County, Maine.\\
        2. The county was incorporated on June 25, 1789 and named for John Hancock, the first governor of the Commonwealth of Massachusetts.\\
        \midrule
        \textbf{\texttt{Target}} \\
        John Hancock\\
    \bottomrule
    \end{tabular}
    \caption{A data example of the reading comprehension task. Data source: HotpotQA.}
    \label{tab:exp11}
\end{table*}

\begin{table*}[t]
    \centering
    \small
    \begin{tabular}{p{.95\linewidth}}
    \toprule
        \textbf{Reranking} \\
    \midrule
        \textbf{\texttt{Input}} \\
        In the reranking task, search engines must understand the relationship between the user's query, which may be keywords or a sentence, and the potential documents. The goal is to ensure that the most relevant documents, those that best cover the user's information needs, are ranked highest. This requires a nuanced understanding of both the query's intent and the content of the documents.\\
        \\
        Investigate the relationship between the document:\\
        What's the point of your question? Are looking for the alcoholic drinks that will least interfere with your fat loss goals or make you pack on the least fat? Straight distilled liquor has zero carbs, all the calories are from alcohol. Drink with ...\\
        and query - what alcohol has the least amount of carbs. Ascertain the document's relevance to the given query, providing a definitive response of `Yes' if the document is relevant to the query or `No' if not.\\
        \midrule
        \textbf{\texttt{Target}} \\
        Yes\\
    \bottomrule
    \end{tabular}
    \caption{A data example of the reranking task. Data source: MS MARCO.}
    \label{tab:exp12}
\end{table*}

\end{document}